\documentclass[lettersize,journal]{IEEEtran}
\usepackage{amsmath,amsfonts}
\usepackage{algorithmic}
\usepackage{array}
\usepackage[caption=false,font=normalsize,labelfont=sf,textfont=sf]{subfig}
\usepackage{textcomp}
\usepackage{stfloats}
\usepackage{url}
\usepackage{color}
\usepackage{verbatim}
\usepackage{graphicx}
\usepackage{cite}
\usepackage{xcolor}
\usepackage{xpatch}
\usepackage[section]{placeins}
\usepackage{booktabs}
\usepackage[ruled]{algorithm2e} % 编写伪代码算法
\makeatletter
\renewcommand{\maketag@@@}[1]{\hbox{\m@th\normalsize\normalfont#1}}%
\makeatother
\def\BibTeX{{\rm B\kern-.05em{\sc i\kern-.025em b}\kern-.08em
		T\kern-.1667em\lower.7ex\hbox{E}\kern-.125emX}}
\usepackage{balance}

\begin{document}
\title{Heterogeneous Network Based Contrastive Learning Method for PolSAR Land Cover Classification}
\author{Jianfeng Cai,Yue Ma,~\IEEEmembership{Member,~IEEE}, Zhixi Feng,~\IEEEmembership{Member,~IEEE},  Shuyuan Yang,~\IEEEmembership{Senior Member,~IEEE}
\thanks{This work was supported by the National Natural Science Foundation of China (Nos. 62171357, 62276205);  the Foundation of Intelligent Decision and Cognitive Innovation Center of State Administration of Science, Technology and Industry for National Defense, China. \textit{(Corresponding author:Yue Ma, Shuyuan Yang.)}

All the authors are with the School of Artificial Intelligence, Xidian University, Xi’an 710071, China (e-mail: jfcai\_1@stu.xidian.edu.cn; mayuedora7@xidian.edu.cn; zxfeng@xidian.edu.cn;  syyang@xidian.edu.cn).}}

\markboth{IEEE Journal of Selected Topics in Applied Earth Observations and Remote Sensing}%
{Contrastive Learning-Based Heterogeneous Network for PolSAR Land Cover Classification}

\maketitle

\begin{abstract}
Polarimetric synthetic aperture radar (PolSAR) image interpretation is widely used in various fields. Recently, deep learning has made significant progress in PolSAR image classification. Supervised learning (SL) requires a large amount of labeled PolSAR data with high quality to achieve better performance, however, manually labeled data is insufficient. This causes the SL to fail into overfitting and degrades its generalization performance. Furthermore, the \emph{scattering confusion} problem is also a significant challenge that attracts more attention. To solve these problems, this article proposes a Heterogeneous Network based Contrastive Learning method(HCLNet). It aims to learn high-level representation from unlabeled PolSAR data for few-shot classification according to multi-features and superpixels. Beyond the conventional CL, HCLNet introduces the heterogeneous architecture for the first time to utilize heterogeneous PolSAR features better. And it develops two easy-to-use plugins to narrow the domain gap between optics and PolSAR, including feature filter and superpixel-based instance discrimination, which the former is used to enhance the complementarity of multi-features, and the latter is used to increase the diversity of negative samples. Experiments demonstrate the superiority of HCLNet on three widely used PolSAR benchmark datasets compared with state-of-the-art methods. Ablation studies also verify the importance of each component. Besides, this work has implications for how to efficiently utilize the multi-features of PolSAR data to learn better high-level representation in CL and how to construct networks suitable for PolSAR data better.
\end{abstract}


\section{Introduction}
\label{section:1}
Polarimetric synthetic aperture radar (PolSAR), an active remote sensing technology, has attracted significant attention due to its ability to obtain richer information than conventional single-polarization synthetic aperture radar (SAR). By using different polarimetric combinations of transmitting and receiving backscattering waves from land covers, PolSAR can observe targets in all-weather and all-time. Therefore, PolSAR image classification \cite{9064495} \cite{8784406}, which is the most crucial task in PolSAR image interpretation, has been widely used in various fields such as geography \cite{hong2015fully}, agriculture \cite{ulaby1990radar}, and environmental monitoring \cite{shirvany2012ship}.

\begin{figure}[!t]
	\centering
	\setlength{\abovecaptionskip}{0mm}
	\setlength{\belowcaptionskip}{-3mm}
	\includegraphics[width=3.5in]{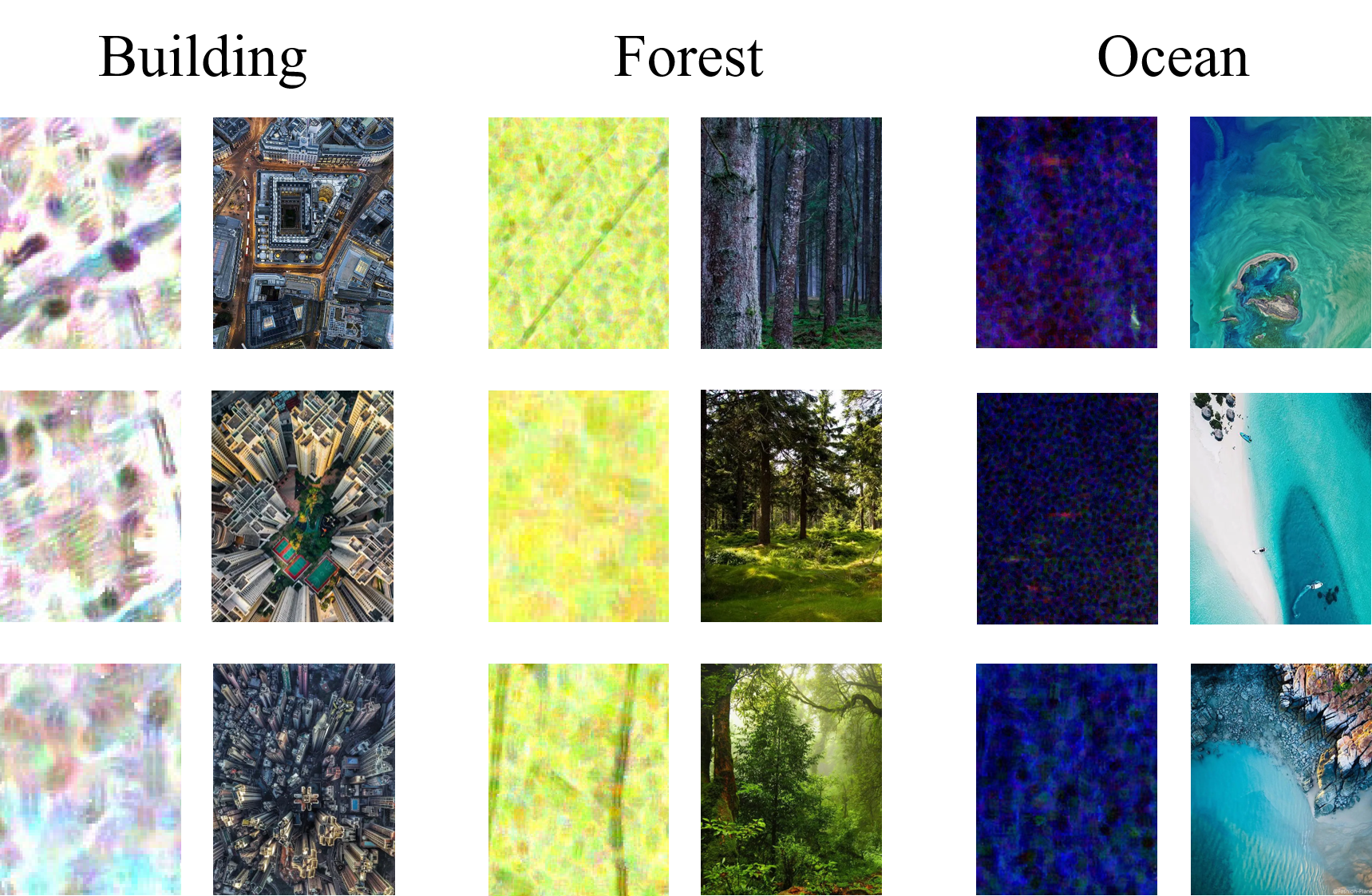}
	\captionsetup{font={small}}
	\caption{Visual comparison of instance similarity between PolSAR and optical images, with PolSAR images on the left and optical images on the right.}
	\label{fig:fig1}
\end{figure}

Numerous researchers have proposed PolSAR classification methods using hand-crafted features. Two primary categories for classifying these features are inherent physical scattering and statistical features. The former is mainly based on target decomposition mechanisms: Freeman decomposition \cite{freeman1998three} decomposes the pixel into three scattering categories; H/A/$\alpha$ decomposition \cite{cloude1997entropy} obtains entropy, anisotropy, alpha angle, and other decomposition methods, including Pauli decomposition \cite{pottier1993dr}, Huynen decomposition \cite{huynen1993physical}, Cameron decomposition \cite{cameron1990feature}, Krogager decomposition \cite{krogager1990new} and so on. The latter mainly consists of the coherency and covariance matrix, which follow the complex Wishart distribution. These methods use classifiers such as SVM \cite{lardeux2009support} and MLP \cite{zou2010polsar} to classify PolSAR data. However, the performance of these methods is heavily dependent on the quality of the features, and none of them can fully represent PolSAR data.

Recently, deep learning (DL) methods, especially convolutional neural networks (CNNs), have achieved magnificent success in various fields, including optical image \cite{krizhevsky2017imagenet, szegedy2015going, he2016deep}, natural language processing (NLP) \cite{zaremba2014recurrent, greff2016lstm, chung2014empirical}, and remote sensing image \cite{9476032, 10341254, ma2019deep, 9970571, 10363338}. Due to the remarkable results of DL, many researchers have proposed several PolSAR deep learning methods. Zhou et al. \cite{zhou2016polarimetric} first used a CNN to replace conventional methods and achieved breakthrough results. To better adapt to the data structure of PolSAR, complex-valued CNNs (CV-CNNs) \cite{zhang2017complex} were proposed. Subsequently, Wishart deep belief networks (WDBNs) \cite{liu2016pol}, fully convolutional networks \cite{mullissa2019polsarnet}, and 3D convolution-based networks \cite{tan2019complex} have been proposed for the characteristics of PolSAR. Specifically, SPAM-Net \cite{9064495} proposed a CNN-based SAR target recognition network that marginalizes the conditional probabilities of SAR targets over their pose angles to precisely estimate the true class probabilities. MvRLNet \cite{9970571} used a novel multi-view GCN-based representation learning network to improve the classification. While supervised CNN-based methods have achieved promising performance, they require a large labeled training set, which is a significant expense of time and energy. When labeled data are scarce, the trained network can easily result in overfitting, leading to a lack of generalization. It means that supervised learning methods lack robustness in the case of missing labeled data, even with augmentation and regularization techniques \cite{srivastava2014dropout, nair2010rectified, srivastava2015highway}.

In contrast to supervised learning, self-supervised learning (SSL) \cite{jing2020self}, where the data provides supervision, has the advantage of learning general representation from unlabeled data, which is more desirable and meaningful. And Contrastive learning (CL), which constructs simple and easy-to-use frameworks for training, is the popular SSL method in optical images recently. From InstDisc \cite{wu2018unsupervised}, CPC \cite{oord2018representation} to MoCo \cite{he2020momentum} and SimSiam \cite{chen2021exploring}, CL has a relatively mature architecture. Moreover, ICDC \cite{9761901} presented a novel two-stage framework that combines Instance-CL and unsupervised clustering to progressively learn desirable temporal representations with high intra-class compactness. \cite{9674754} proposed Inter-Intra Contrastive (IIC) framework to learn video feature representations. Consistent intra-VCL \cite{9893855} introduced a novel intra-video contrastive learning (intra-VCL) that further distinguishes intra-video actions. However, the gap between PolSAR and optical images/videos makes directly applying optical CL methods to PolSAR tricky. Therefore, Researchers have proposed some PolSAR-tailored CL methods to address this issue: MFM \cite{10493132} proposed masked feature modeling, a methodology for the generative selfsupervised learning of high-resolution remote sensing images that combines convolutional neural network (CNN) and Transformer architectures. Li et al. \cite{9460820}  designed three different pretext tasks and a triplet Siamese network to learn the high-level and low-level image features at the same time. MI-SSL \cite{ren2021mutual} learned the implicit multi-modal representation from unlabeled data. Yan et al. \cite{10109790} proposed a novel domain knowledge-guided self-supervised learning approach for unsupervised CD by fusing the domain knowledge of remote sensing indices during training and inference. PCLNet \cite{zhang2020unsupervised} developed an instance discrimination proxy objective to learn representation from unlabeled data. SSPRL \cite{zhang2022exploring} improves CL, so no negative samples are needed; TOV \cite{10110958} indicated that The Original Vision can be easily adapted to various tasks tTrained by massive unlabeled optical data along a human-like self-supervised learning (SSL) path that is from general knowledge to specialized knowledge. Cui et al. \cite{cui2022tcspanet} proposed TCSPANet, the two-staged CL based on attention. However, there are still some challenges with these methods:

\begin{itemize}
	\item Many of them fail to utilize the multi-features in PolSAR data fully. The diversity of PolSAR features makes it more advantageous in CL, which can extract high-level representation more easily. However, similar to the standard architecture for CL of optical images, they directly utilize the Siamese network to PolSAR, making it hard to exploit these features thoroughly. On the contrary, the heterogeneous network can arbitrarily combine different features for better representation learning.
	
	\item Some methods attempt to incorporate different features but fail to consider the redundancy between them. The degree of information redundancy varies for different feature combinations, and some may even hinder model learning. Therefore, feature selection is essential.
	
	\item Some of them ignored the high similarity between pixels in PolSAR data. As shown in Fig.\ref{fig:fig1}, instances in optical images are generally images with thousands of pixels, and even images of the same class have very different pixel values among themselves. However, in PolSAR, the instance is only the pixel with the scattering matrix and even the pixels of different classes have relatively high similarities. This makes the contrastive learning framework which outperforms in the optical image, struggle to learn the discriminative representation in PolSAR. To overcome this limitation, introducing diversity between instances is necessary to enhance model learning. 
	
	\item All of these methods do not consider the scattering confusion problem. Due to the scattering mechanism, there are many land covers whose scattering information is strongly correlated with each other, resulting in the model cannot classify them well. To solve this problem, the model needs to learn the scattering difference among different land covers using a large number of instances with scattering features.
\end{itemize}

Based on the above analysis, this article introduces a novel approach for PolSAR classification, named Heterogeneous Contrastive Learning Network (HCLNet). The proposed HCLNet employs two types of features, physical and statistical, for CL. For the physical features, it uses the feature filter to reduce feature redundancy, while for the statistical features, it utilizes the coherency matrix directly. Additionally, it constructs a heterogeneous network to learn the representation of PolSAR data using the novel superpixel-based Instance Discrimination. This approach effectively utilizes the PolSAR multi-features and addresses the problem of pixel similarity and scattering confusion. Furthermore, it uses two easy-to-use plugins to better adapt to PolSAR. Specifically, the main contributions of this work can be summarized as follows: 

\begin{itemize}
	\item A \emph{Heterogeneous Network} is proposed to learn the representation hidden among PolSAR data in CL for the first time to effectively alleviate the challenge of scattering confusion. The network consists of two sub-networks with \emph{heterogeneous} architectures where the former is a 2D CNN, and the latter is a 1D CNN. This network can input multi-features, including physical and statistical features, to construct \emph{heterogeneous} features for extracting high-level representations in an unsupervised paradigm.
	
	\item A novel pretext task, \emph{Superpixel-based Instance Discrimination}, is designed to reduce the similarity between pixels and thus the model can learn representation easier and better. This task utilizes superpixel segmentation to select positive and negative samples for CL, which reduces the occurrence of highly similar pixels being negative samples of each other. 
	
	\item \emph{Feature Filter} is introduced to select complementary features and reduce redundancy. It created a classifier as the criterion to implement a suitable combination of these features and remove redundant information from multi-features.
	
	\item HCLNet is evaluated on three benchmark PolSAR classification datasets, and the experimental results demonstrate the superiority over its counterparts.
\end{itemize}

The rest of this article is organized as follows: A brief review of the CL method for optical images and PolSAR and the introduction of PolSAR multi-features are given in Section \ref{section:2}. The details of the HCLNet are described in Section \ref{section:3}. And Section \ref{section:4} shows the experimental results and analysis. Finally, we provide the conclusion and prospects for future research in Section \ref{section:5}.

\section{Related Work}
\label{section:2}
\subsection{Contrastive Learning}
\label{section:2.1}
As a prevalent SSL method, CL has a mature and general architecture. It usually has two networks with the same architecture, which also can share parameters: one is called the online network, which will be sent to downstream tasks for fine-tuning as the main network, and the other is called the target network to auxiliary learning. Generally, CL is trained by a proxy objective, called pretext task, which usually chooses Instance Discrimination with InfoNCE loss function. Its main idea is that two representation obtained by the two networks from different perspectives of the same image should be similar and vice versa. And InfoNCE loss function is a modified Cross-Entropy Loss that introduces the dot product to compute similarity. The formula is as follows:

\begin{equation}
	L(q, k) = - log \frac{exp(q \times k_{+} / \tau)}{\sum_{i=0}^{K}{exp(q \times k_{i} / \tau)}}
\end{equation}

{
	\setlength{\parindent}{0cm}where $q$ is the output of the online network, $k_{+}$ is the output of the target network that $q$ matches and is called the positive sample, $k_{i}$ ($i=0 \cdots K$ and $k_{i} \neq k_{+}$) is all output of the target network, which is memoried and is called the negative sample, $\tau$ is a temperature hyper-parameter that adjusts the uniformity of information distribution \cite{wu2018unsupervised}. Intuitively, this loss tries to classify $q$ to $k_{+}$, and essentially, it is the log loss of a ($K+1$)-way softmax-based classifier. 
}
\subsubsection{The CL in Optical Images}
\label{section:2.1.1}
In optical images, according to the different ways of parameter updating and negative sample selection, CL methods can be divided into different types. Wu et al. \cite{wu2018unsupervised} first proposed Instance Discrimination with NCE loss function and memory bank to store negative samples. It treats the two sub-images obtained from a cropped image as the positive sample and all other images in the dataset as negative samples. While Ye et al. \cite{ye2019unsupervised} selected other samples in the same mini-batch as negative samples. CPC \cite{oord2018representation} is a more general architecture containing an encoder and an auto-regression model. Positive and negative samples are constructed to train the encoder autoregressively. In addition to cropping, optical images have properties such as depth that can also form positive samples with each other. So Tian et al. \cite{tian2020contrastive} proposed CMC  using luminance (L channel), chrominance (ab channel) \cite{zhang2017split}, depth, surface normal \cite{eigen2015predicting}, and semantic labels to construct the positive sample. In order to retain more negative samples and smooth change sample representation, MoCo \cite{he2020momentum} introduced a queue dictionary to increase negative samples and exponential moving average (EMA) to update the parameter of the target network. SimCLR \cite{chen2020simple} used a larger batch size to achieve better results. It also adds a projection head at the end of the network and achieves incredible performance gains. To eliminate negative samples, BYOL \cite{grill2020bootstrap} added a prediction head to the end of the online network, then turns the similarity problem into a prediction problem, which can effectively prevent the model collapse. By summarizing previous works, Chen et al. \cite{chen2021exploring} proposed a simple architecture SimSiam and demonstrates the importance of stopping gradients.

\subsubsection{The CL in PolSAR}
\label{section:2.1.2}
To use CL methods in the PolSAR domain, some researchers proposed PolSAR-tailored CL methods for improvement: MI-SSL \cite{ren2021mutual} uses coherency matrix $T$ and constructs positive samples by visual, physical, and statistical features to learn the implicit multi-modal representations with similarity and difference loss; PCLNet \cite{zhang2020unsupervised} which copies the MoCo \cite{he2020momentum} technique, selects dataset according to stimulation for interclass and intraclass diversity and uses PolSAR image rotating $180^\circ$ as the positive sample. SSPRL \cite{zhang2022exploring} proposed two branches and dynamic convolution (DyConv) layer to improve CL, and its basic architecture is similar to BYOL. TCSPANet \cite{cui2022tcspanet} exploited unsupervised multi-scaled patch-level datasets (UsMsPD) and semi-supervised multi-scaled patch-level datasets (SsMsPD). It also proposes two CL stages in TCNet and adds attention mechanism in SPAE to get better results. 

In this article, inspired by SimCLR, we construct a heterogeneous CL network based on superpixel-based instance discrimination, which selects appropriate negative samples to reduce the high similarity between positive and negative samples. 

\subsection{Multi-features within PolSAR} 
\label{section:2.2}
\subsubsection{Physical scattering features}
\label{section:2.2.1}
As mentioned in \cite{yang2019cnn}, most physical scattering features are based on target decompositions. Different target decompositions, which decompose targets based on the scattering matrix $S$, have distinct advantages for PolSAR image classification. For example, Freeman decomposition \cite{freeman1998three} decomposes the scattering matrix $S$ into three scattering categories: surface, volume, and double bounce; entropy, anisotropy, and alpha angle are obtained by H/A/$\alpha$ decomposition \cite{pottier1993dr}; Krogager decomposition \cite{krogager1990new} decomposes the scattering matrix $S$ into sphere, diplane, and helix components. Other decomposition methods include Yamaguchi, Vanzyl, Neuman, Multiple-Component Scattering Model (MCSM), Huynen, Holm, Barnes, Cloude, Anned, An-Yang, Pauli decomposition, Huynen decomposition, Cameron decomposition \cite{pottier1993dr} \cite{huynen1993physical} \cite{cameron1990feature} \cite{cloude1996review, yamaguchi2005four, van1993application, zhang2008multiple, holm1988radar} and so on, which is shown in Table \ref{table:tab1}.

These features have different importances in specific scenes and some features have functional overlaps, which leads to feature redundancy. So combining these features properly is necessary to learn representation better. We propose the feature filter to obtain a better complementary feature combination that is similar to \cite{yang2019cnn}. 

\subsubsection{Statistical features}
\label{section:2.2.2}
According to the statistical characteristics, PolSAR data can also become the coherency matrix $T$ and the covariance matrix $C$ based on the scattering matrix $S$, which follows the complex Wishart distance. Specifically, the scattering matrix $S$ is defined as

\begin{equation}
	S=\left[\begin{array}{ll}
		S_{H H} & S_{H V} \\
		S_{V H} & S_{V V}
	\end{array}\right]
\end{equation}

{
	\setlength{\parindent}{0cm}where $S_{X Y}$,$X$,$Y\in[H,V]$ is the scattering element of horizontal/vertical transmitting/receiving polarization. Then the covariance matrix $C$ is formed by
}

\begin{equation}
	h=\left[\begin{array}{lll}
		S_{H H} & \sqrt{2} S_{H V} & S_{V V}
	\end{array}\right]^T
\end{equation}

\begin{scriptsize}
	\begin{equation}
		\hspace{-1mm}
		C=h h^{*T}=\left[\begin{array}{ccc}
			\left|S_{H H}\right|^2 & \sqrt{2} S_{H H} S_{H V}{ }^* & S_{H H} S_{V V}{ }^* \\
			\sqrt{2} S_{H V} S_{H H}{ }^* & 2\left|S_{H V}\right|^2 & \sqrt{2} S_{H V} S_{V V}{ }^* \\
			S_{V V} S_{H H}{ }^* & \sqrt{2} S_{V V} S_{H V}{ }^* & \left|S_{H V}\right|^2
		\end{array}\right]
	\end{equation}
\end{scriptsize}

{
	\setlength{\parindent}{0cm}where the superscript “$*T$” denotes the conjugate transpose. The coherent matrix $T$ is formed by
}

\begin{footnotesize}
	\begin{equation}
		\hspace{-1mm}
		k_p=\left[\begin{array}{llll}
			\left(S_{H H}+S_{V V}\right) / \sqrt{2} & \left(S_{H H}-S_{V V}\right) / \sqrt{2} & \sqrt{2} S_{H V}
		\end{array}\right]^T
	\end{equation}
\end{footnotesize}

\begin{equation}
	T=<k_p k_p^{*T}>=\left[\begin{array}{lll}
		T_{11} & T_{12} & T_{13} \\
		T_{21} & T_{22} & T_{23} \\
		T_{31} & T_{32} & T_{33}
	\end{array}\right]
\end{equation}

\subsubsection{Feature Selection}
\label{section:2.2.3}
Feature selection is an essential problem in PolSAR. There are different levels of complementarity and redundancy between different polarized features, and a good feature selection method can well preserve the complementarity and eliminate the redundancy by the combination of multi-features. Haddadi G et al. \cite{haddadi2011polarimetric} proposed the PolSAR feature selection method using a combination of a genetic algorithm (GA) and am artificial neural network (ANN). Yang et al. \cite{yang2019cnn} used the Kullback-Leibler distance (KLD) as a criterion, and a 1-D CNN as a selection model to select feature subsets given the number of features. Huang et al. \cite{huang2021multi} realized multi-view feature selection via manifold regularization and $l_{2,1}$ sparsity regularization, including polarimetric features and texture features. In addition, many studies also introduced the attention mechanism into feature selection. Specifically, AFS-CNN \cite{dong2020attention} is proposed to capture the relationship between input polarimetric features through attention-based architecture to ensure the validity of high-dimensional data classification. In contrast, our feature selection method, called feature filter, is similar to \cite{yang2019cnn}, but uses a simpler and more intuitive feature selection objective to obtain the better combination of features.

\section{Heterogeneous Network based Contrastive Learning}
\label{section:3}
\begin{figure*}[!t]
	%	\vspace{1mm}
	\centering
	\setlength{\abovecaptionskip}{-3mm}
	\setlength{\belowcaptionskip}{-3mm}
	\includegraphics[width=7in]{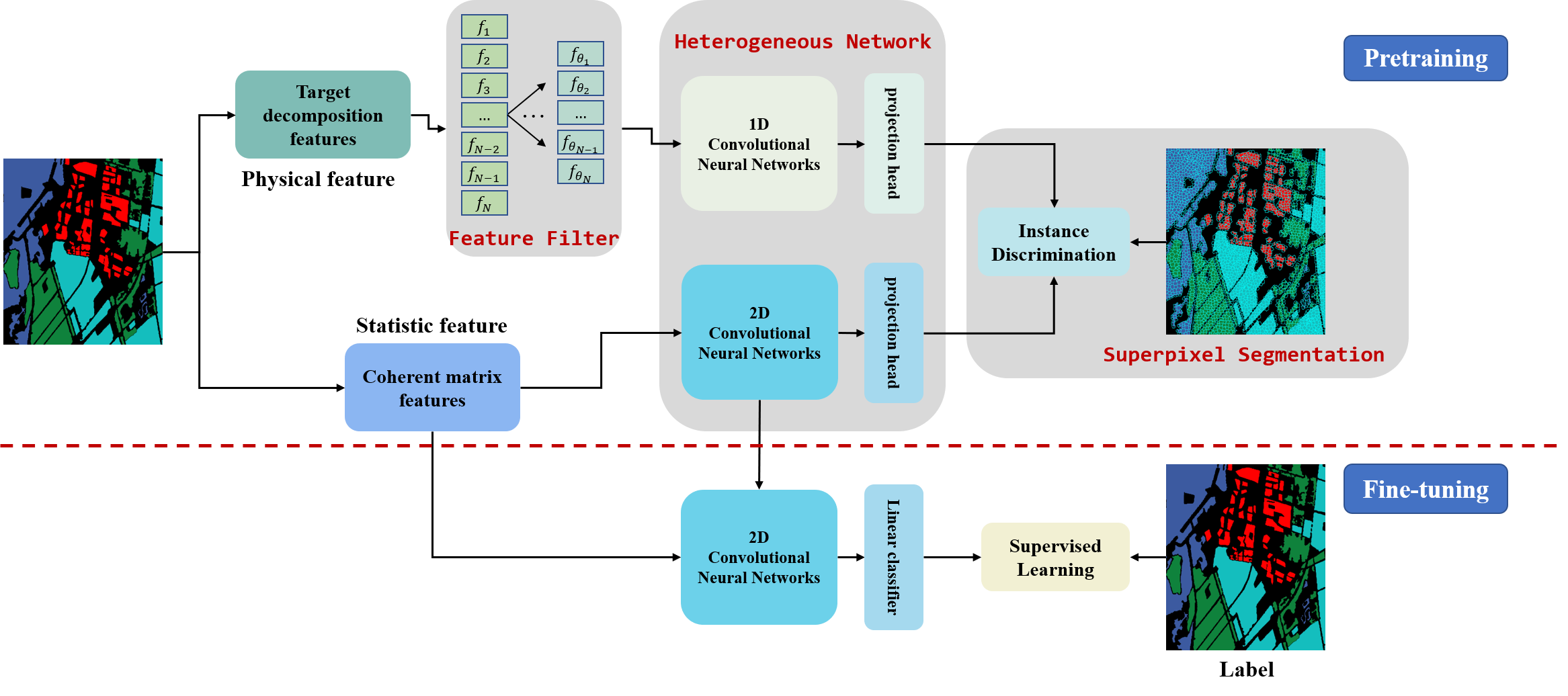}
	\captionsetup{font={small}}
	\caption{The overall framework of the proposed HCLNet. It mainly contains two processes: Pretraining and Fine-tuning. In pretraining, it first uses Feature Filter to combinate features, then constructs the heterogeneous network and uses Superpixel-based Instance Discrimination to learn the high-level representation. In fine-tuning, it uses the trained online network from pretraining and fine-tunes it with a small number of labeled data to better fit the downstream distribution.}
	\label{fig:fig2}
\end{figure*}

\begin{figure}[!t]
	\setlength{\abovecaptionskip}{-3mm}
	\setlength{\belowcaptionskip}{-3mm}
	\includegraphics[width=3.5in]{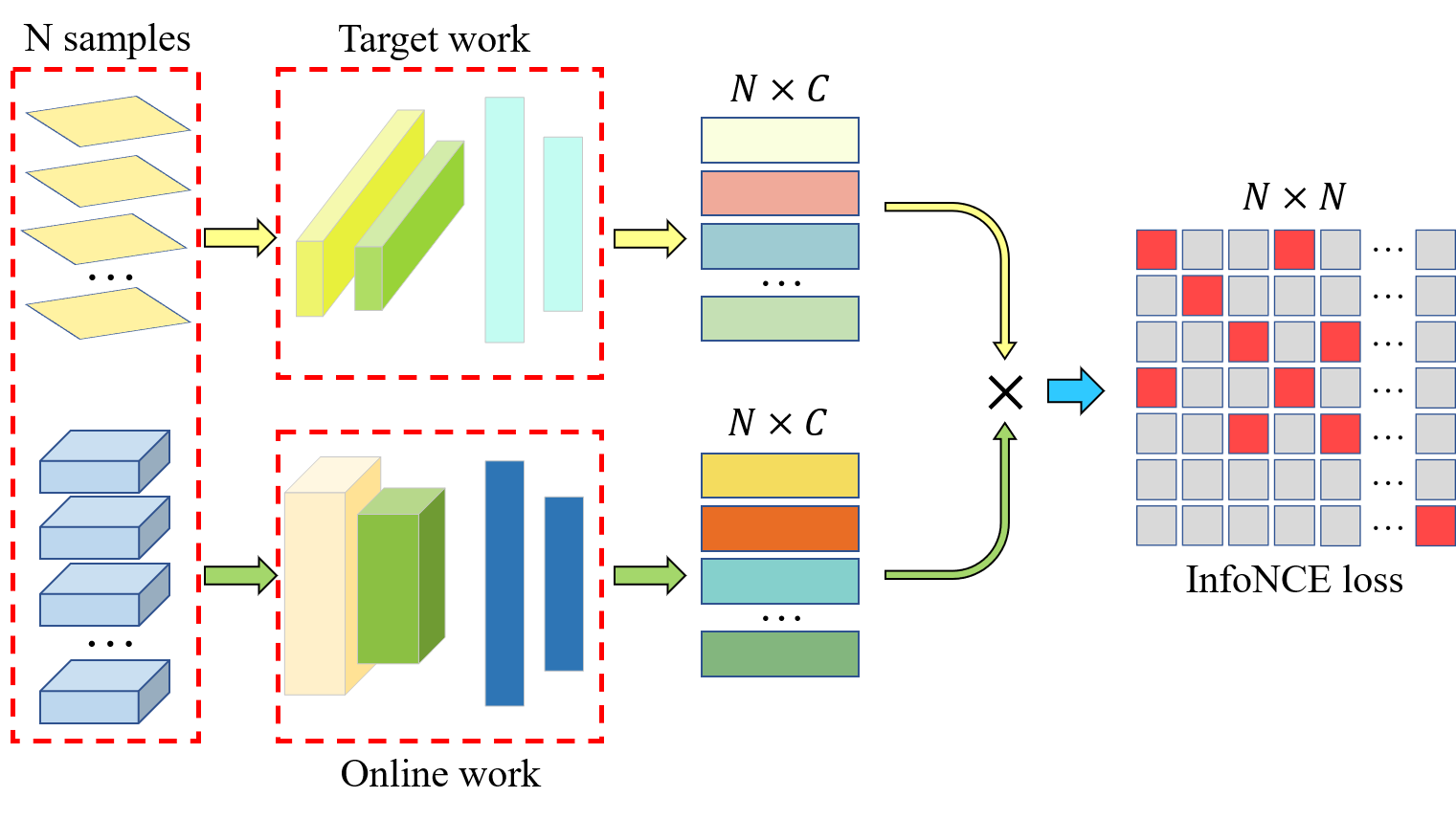}
	\captionsetup{font={small}}
	\caption{The architecture of the heterogeneous network in HCLNet. It contains two networks with different architectures and is updated with InfoNCE loss. The output of the target network belonging to different superpixels in the same minibatch will be served as negative samples.}
	\label{fig:fig3}
\end{figure}

\begin{figure}[!t]
	\setlength{\abovecaptionskip}{-3mm}
	\setlength{\belowcaptionskip}{-3mm}
	\includegraphics[width=3.5in]{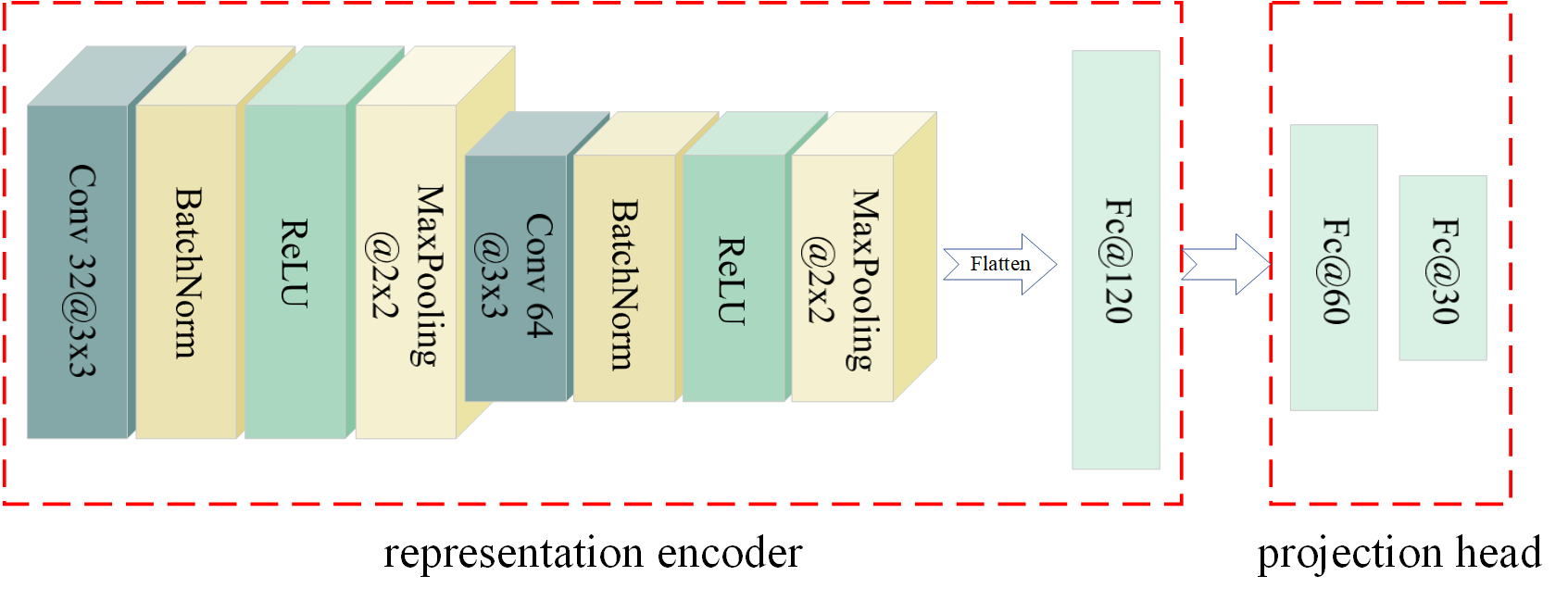}
	\captionsetup{font={small}}
	\caption{The architecture of the online network in the heterogeneous network. It contains the representation encoder and the projection head; the former will be used for fine-tuning.}
	\label{fig:fig4}
\end{figure}

In this section, we present the detailed process of the proposed HCLNet. The overall architecture of HCLNet is shown in Fig.\ref{fig:fig2}, which includes three main components. Among them, the first component is the feature filter, which finds the appropriate combination of target decomposition features(physical features). The second component is Superpixel-based Instacne Discrimination, which improves the selection of positive and negative samples using superpixel in Instance Discrimination. Moreover, the final component is the Heterogeneous Network, which is the most important one and learns the high-level representation of PolSAR data using different network architectures. First, all target decomposition features are filtered using the feature filter to obtain the complementary feature combinations. Then the coherent matrix and the filtered target decomposition features are used as the input of the heterogeneous network to learn the high-level representation of PolSAR instances. And the superpixel-based instance discrimination is used for unsupervised training. Finally, the pre-trained online network is used for the downstream task by fine-tuning. The specific details of each component are as follows. 

\subsection{Feature Filter}
\label{section:3.1}
Here, we will introduce the feature filter in detail. Like \cite{yang2019cnn}, we extract multi-features by $M$ target decomposition methods mentioned in Section \ref{section:2.2}, which have $N$ in total, and $M$ groups, because each target decomposition method generates a group of features. Then, we design a 1-D CNN model as the classifier namely $B_c$ to evaluate the performance of different combinations of features. The network's input is an $N{\times}1$ vector $V$, representing the $N$ features of a pixel. During training, we use all $N$ features to predict the label of each pixel for supervised learning with least-squares loss function \cite{holm1988radar}. 

After training, we use beam search to select the appropriate combination of features and use classifier accuracy as the selection criterion. Different from \cite{yang2019cnn}, we do not introduce the additional KLD criterion to interfere with the 1D-CNN classifier’s choice of features combinations. In this way, the complementarity of the feature groups selected by the classifier can be guaranteed to the greatest extent and the redundancy between features can be eliminated as much as possible. Specifically, Step 1, we start with the initial $M$ group of features and choose to remove the first $k$ groups of features that cause the slightest reduction in classification accuracy to form $k$ branches, where one group of features is removed from each branch. Step 2, in each branch, the above steps are repeated such that each branch forms $k$ branches, and the total branches are $k{\times}k$. Step 3, we choose the first $k$ branches, according to the classification accuracy from high to low. Then repeating Step 2 and Step 3 until the feature groups number is reduced to the threshold $\theta$. The process of selecting features by the feature filter is outlined in Algorithm 1. 

\begin{algorithm}[htb]
	\setlength{\abovecaptionskip}{-7mm}
	\setlength{\belowcaptionskip}{-7mm}
	\caption{Feature filter for selecting the appropriate combination of feature groups}
	%\LinesNumbered % 要求显示行号
	\KwIn{feature groups set $M_i$, the number of features $N$, group threshold $\theta$, branch number $k$, classifier $F$} % 输入参数
	\KwOut{selected feature groups set $M_o$} % 输出
	
	set $Q$ = $\{M_i\}$\
	
	\While{$N$ $>$ $\theta$}
	{
		$Q'$ $\leftarrow$ $\{\}$
		
		\For{feature groups $M$ in $Q$}
		{
			\For{feature $f$ in $M$}
			{
				remove $f$ from $M$
				
				\eIf{len($Q'$) == $k$}
				{
					\If{$F$($M$) $\ge$ max($F$($Q'$))}
					{
						pop $Q'$(max($F$($Q'$)))
						
						push $M$ into $Q'$
					}
				}
				{
					push $M$ into $Q'$
				}
			}
		}
		N $\leftarrow$ N-1
		
		$Q \leftarrow Q'$
	}
\end{algorithm}

To ensure the unity of the input dimensions of the classifier in each iteration, we directly set the value of the features removed each time to $0$. Finally, we obtain $\theta$ group, $\theta_N$ features as the input of the target network, as described in Section \ref{section:3.3}. 

\subsection{Superpixel-based Instance Discrimination}
\label{section:3.2}
As mentioned in Section \ref{section:2.1}, the general pretext task of the optical images is Instance Discrimination with InfoNCE loss function. The input of CL is usually the whole optical image, with both pixel and semantic features. Therefore significant differences exist between different optical images, even in the same class and the common Instance Discrimination, which treats all images other than the current image as negative samples, can perform well. However, PolSAR mainly uses pixels, which has a considerable similarity, as instances for CL training. Therefore, we can no longer treat all other pixels as negative samples of the current pixel. Instead, we should add some prior knowledge to promote the model to select the pixels with large similarity difference to the original pixels as negative samples. In this case, superpixel segmentation algorithm is a simple and efficient unsupervised algorithm for measuring the similarity between pixels. So to continue to take advantage of Instance Discrimination, we improve the way of selecting negative samples to select them more reasonably according to the superpixels. The details are as follows: 

First, we segment the PolSAR image into some superpixels, as shown in Fig.\ref{fig:fig5}. Specifically, we choose the classical superpixel segmentation method: Simple Linear Iterative Clustering (SLIC) \cite{achanta2012slic} to segment the whole PolSAR images. Compared to the other algorithm, such as turbo or Ncut, it is fast, memory efficient, boundary adherence, and needs to set a few parameters. It used the idea of clustering to cluster part of the pixels into a superpixel. So the similarity of pixels within the same superpixel is high and vice versa. Then, within the scope of superpixel segmentation, we redefine the positive and negative sample, that is, pixels within different superpixels are defined as negative samples of each other and within the same superpixel as positive samples. And the filtering physical features and statistical features of each pixel remain unchanged. Assumed the size of the PolSAR image is $H{\times}W$, and we obtain $N_{s}$ superpixels, $N_{s_{i}}$ pixels in $i$th superpixel, the pixel in $i$th superpixel has at most $N_{s_{i}}-1$ positive sample and $H{\times}W-N_{s_{i}}$ negative samples. In general, for a instance sample (that is, a pixel) $p_o$ in a batch $B$, there are $N_+$ samples in the same superpixel with $p_o$ and $N_- = B - N_+ - 1$ samples that are not in the same superpixel with $p_o$. Then the $N_+$ samples and $p_o$ are positive samples, and the remaining $N_-$ samples are negative samples of $p_o$. Therefore, its loss function can be rewritten from the form in Section \ref{section:2.1} as:

\begin{multline}
	L_{p_o}(p_o,\{p_1^+,...,p_{N_+}^+\},\{p_1^-,...,p_{N_-}^-\}) = -\frac{1}{B-1} \\
	\times log\frac{\sum_{i=1}^{N_+}{exp(p_o \times p_i^+ / \tau)}}{\sum_{i=1}^{N_+}{exp(p_o \times p_i^+ / \tau)} + \sum_{j=1}^{N_-}{exp(p_o \times p_j^- / \tau)}}
\end{multline}

{
	\setlength{\parindent}{0cm}where $p_i^+$ and $p_j^-$ represent the $i$th positive and the $j$th negative samples, respectively. Since all $B$ samples in the batch are selected in turn, the total cost function becomes:
}

\begin{equation}
	L_{HCLNet} = \sum_{i=1}^{B} \frac{L_{p_i}(p_i,\{p_1^+,...,p_{N_+^i}^+\},\{p_1^-,...,p_{N_-^i}^-\})}{B}
\end{equation}

{
	\setlength{\parindent}{0cm}where $N_+^i$ and $N_-^i$ represent the number of the positive and negative samples of $p_i$ respectively. In terms of implementation, we can sample one pixel in each superpixel to form a batch, so that we can use traditional Instance Discrimination (Equation (1)) to train the Heterogeneous Network described in Section \ref{section:3.3}. 
}

\subsection{Heterogeneous Network}
\label{section:3.3}
Inspired by CLIP \cite{radford2021learning} in which two networks use different architectures to input different features for fusion learning, PolSAR-tailored Heterogeneous Network is proposed in this article, shown in Fig.\ref{fig:fig3}. It is similar to the traditional CL Siamese Network, which has two networks; however, its architecture is different between the two networks. The online network of the heterogeneous network is a 2-D CNN, while the target network of the heterogeneous network is a 1-D CNN. Morover, different from the traditional CL, Heterogeneous Network inputs different and complementary features, which could better learn the scattering difference among different instances to mitigate the scattering confusion problem. The details of the two networks are as follows:

\subsubsection{Online network}
\label{section:3.3.1}
The online network consists of a 2-D CNN called the representation encoder and a MLP namely the projection head, which is similar to the common CL network. Its input is a two-dimensional block of pixels with size $k{\times}k$, each pixel with the feature value of the coherent matrix $T$. Because of its two-dimensional characteristics, it is used to learn the scattering relationship between the current pixel and its neighboring pixels to better learn the scattering similarity. The typical architecture of the representation encoder is shown on the left of Fig.\ref{fig:fig4}, denoted as $f_e(\cdot)$. Then a projection head $g_p(\cdot)$ is followed by $f_e(\cdot)$, which aims to embed the representation into the more high semantic space, which is also a multi-linear embedding layer. So we obtain the final online network $g_p(f_e(\cdot))$.

The input feature of the online network is the coherency matrix, as mentioned in Section \ref{section:2.2}. Specifically, for the pixel $p$, we crop out a pixel block of size $k{\times}k$ with $p$ as the center, and the value of each pixel is represented by the flattened coherency matrix $\hat{T} \in R^{1{\times}9}$. Finally, the input dimension is $k{\times}k{\times}9$.

\subsubsection{Target network}
\label{section:3.3.2}
The overall architecture of the target network is similar to the online network and includes a representation encoder which is 1D-CNN and a projection head, except they change from two dimensions to one. It's also similar in structure to the classifier $B_c$ used in the feature filter, but they serve very different purposes. Its input is the combination of complementary target decomposition multi-features of the current pixel filtered by the feature filter as described in Section \ref{section:3.1}. Specifically, for pixel $p$, We stack all the $\theta_N$ features that represent a $\theta_N{\times}1$ vector as input. By learning the multi-features of different pixels, the model is prompted to learn the scattering difference. At the same time, through the combination of online network, the heterogeneous network can better learn the scattering differences and similarity among different pixels, which greatly alleviates the problem of scattering confusion.

Finally, for pixel $p$, the online network inputs the matrix of dimension $k{\times}k{\times}9$, and outputs the $m{\times}1$ vector $r_o$ as representation; the target network inputs the vector of dimension $\theta_N{\times}1$, and outputs the vector $r_{t+}$ whose dimension is the same as $r_o$. Then we use the InfoNCE loss function $L(r_o,r)$ to compute the similarity, where $r \in \{r_{t+}, r_{t-}\}$, $r_{t-}$ represents the target network outputs of the other pixels which belong to different superpixels in the same batch. After training, we can obtain the final pre-trained online network to transform PolSAR data into a high-level representation. Then it can be used as the backbone network for downstream classification tasks and performs well with fine-tuning. 

\section{Experimental Results and Analysis}
\label{section:4}
\subsection{Datasets Description}
\label{section:4.1}
In this section, we employ three standard PolSAR datasets to verify the superiority of the HCLNet. They include RADARSAT-2 Flevoland, AIRSAR Flevoland, and ESAR Oberpfaffenhofen.

\begin{figure}[!t]
	\centering
	\setlength{\abovecaptionskip}{0mm}
	\setlength{\belowcaptionskip}{-3mm}
	\includegraphics[width=3.5in]{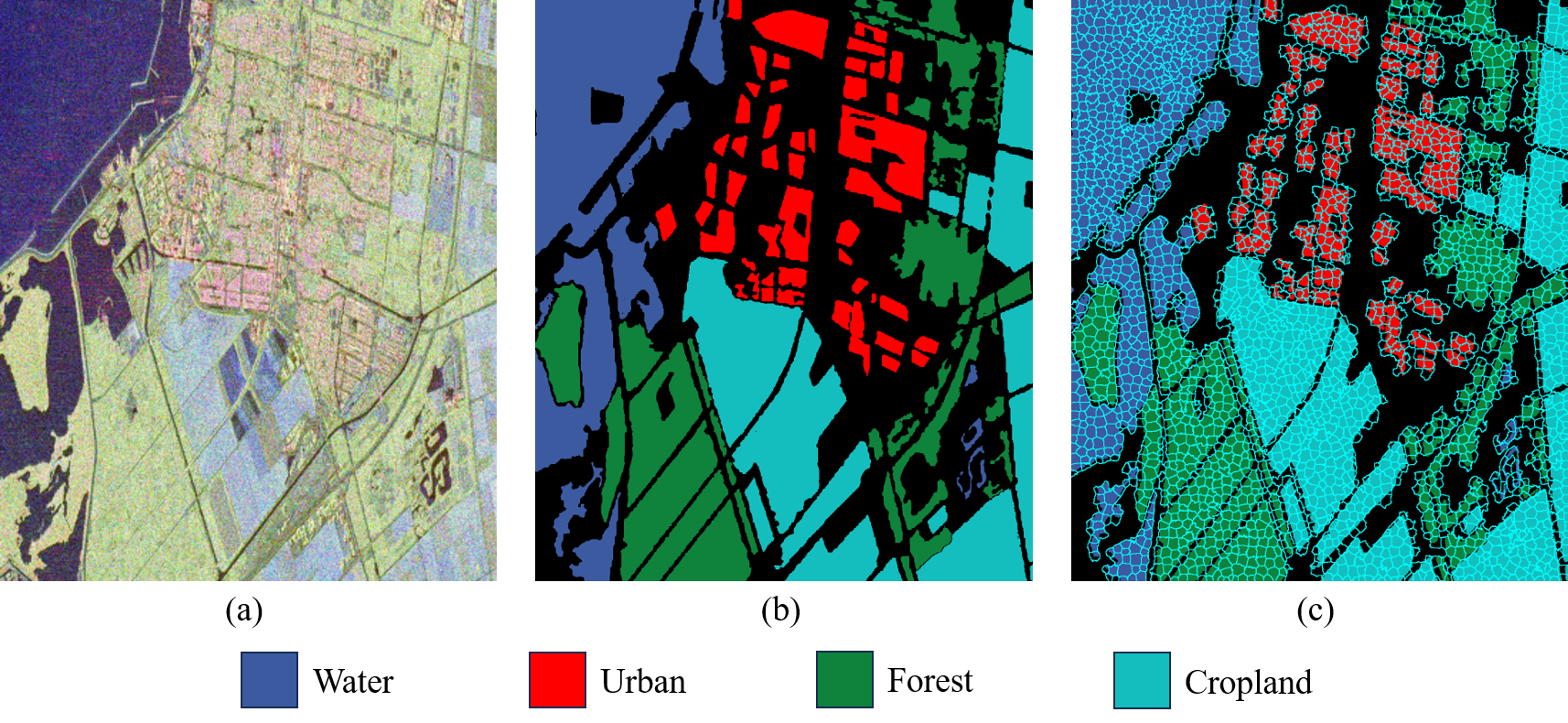}
	\captionsetup{font={small}}
	\caption{RADARSAT-2 Flevoland data. (a) Pauli RGB image. (b) Ground-truth image. (c) Superpixel image.}
	\label{fig:fig5}
\end{figure}

\begin{figure}[!t]
	\centering
	\setlength{\abovecaptionskip}{0mm}
	\setlength{\belowcaptionskip}{-3mm}
	\includegraphics[width=3.5in]{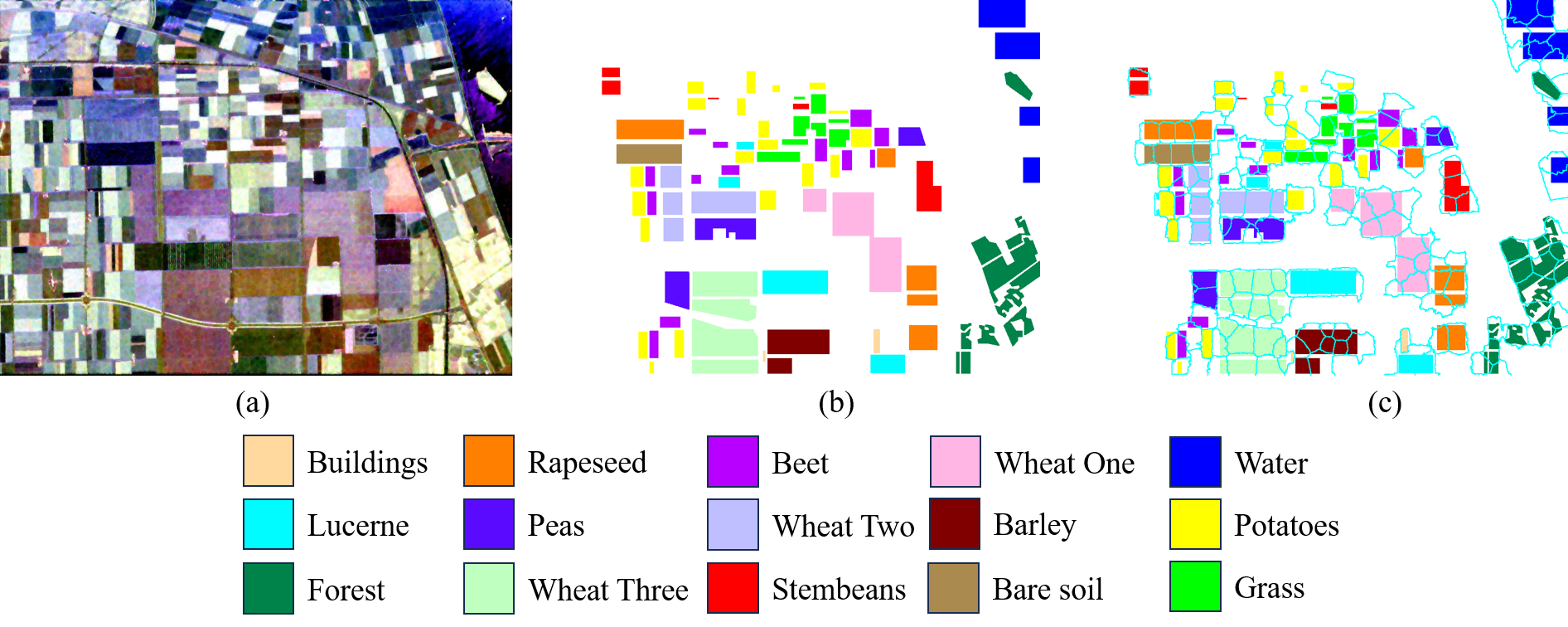}
	\captionsetup{font={small}}
	\caption{AIRSAR Flevoland data. (a) Pauli RGB image. (b) Ground-truth image. (c) Superpixel image.}
	\label{fig:fig6}
\end{figure}

\begin{figure}[!t]
	\centering
	\setlength{\abovecaptionskip}{0mm}
	\setlength{\belowcaptionskip}{-3mm}
	\includegraphics[width=3.5in]{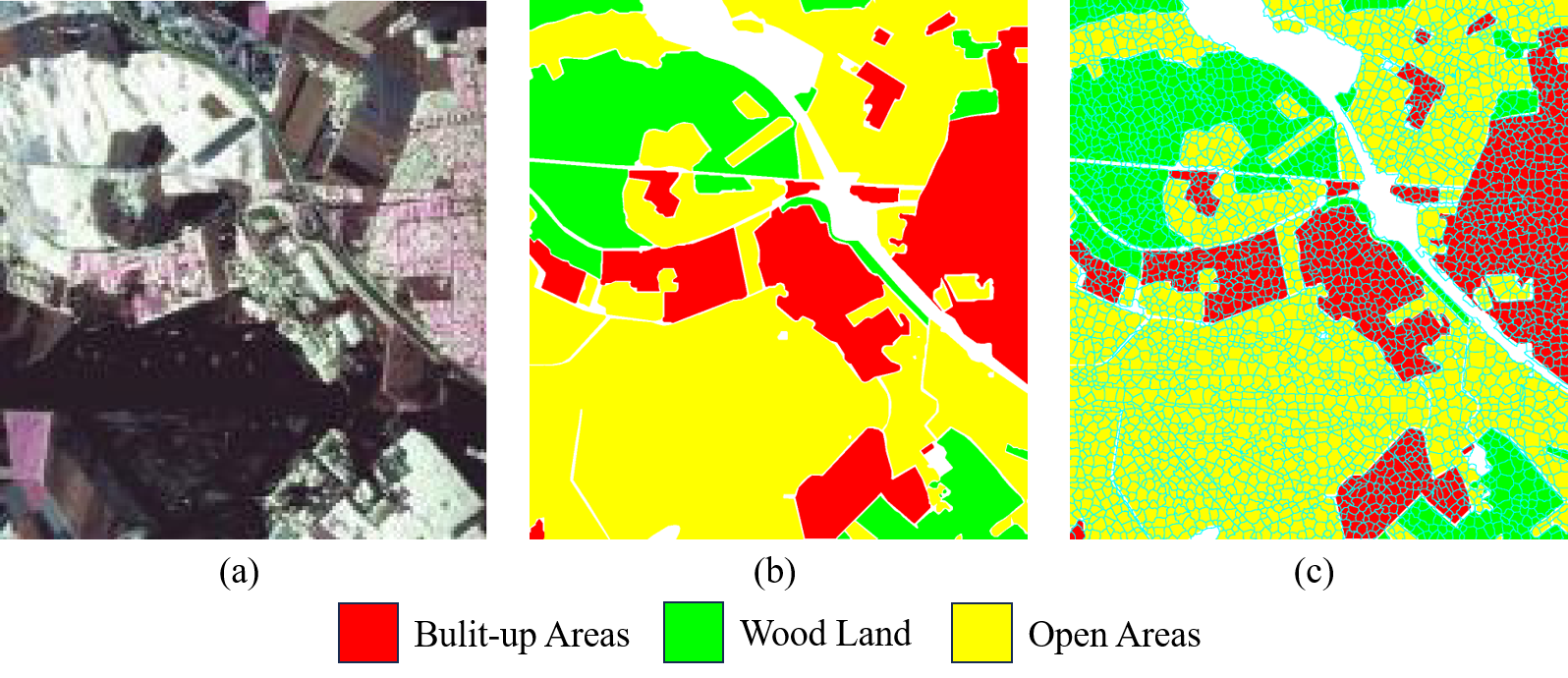}
	\captionsetup{font={small}}
	\caption{ESAR Oberpfaffenhofen data. (a) Pauli RGB image. (b) Ground-truth image. (c) Superpixel image.}
	\label{fig:fig7}
\end{figure}

\begin{figure}[!t]
	\centering
%	\vspace{-3mm}  %调整图片与上文的垂直距离
	\setlength{\abovecaptionskip}{-1mm}
	\setlength{\belowcaptionskip}{-3mm}
	\centering
	\includegraphics[width=3.5in]{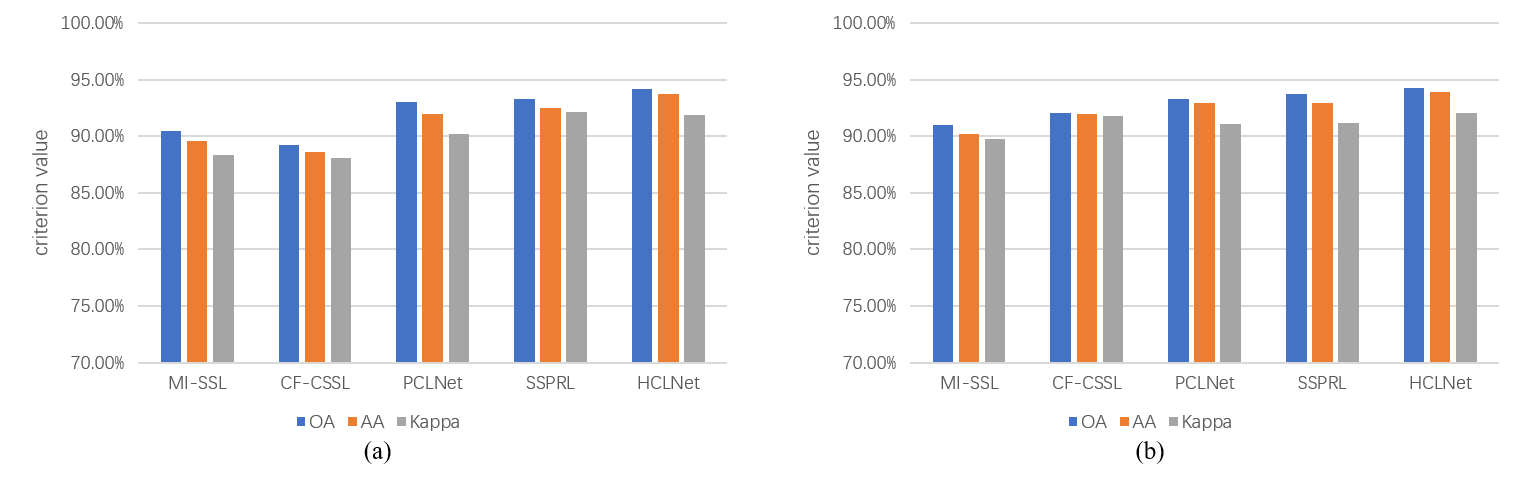}
	\captionsetup{font={small}}
	\caption{Comparisons of different methods on the RADARSAT-2 Flevoland dataset. (a) Few-shot result. (b) Full-sample result.}
	\label{fig:fig8}
\end{figure}

\begin{figure*}[htb]
	\setlength{\abovecaptionskip}{-3mm}
	\setlength{\belowcaptionskip}{-3mm}
	\centering
	\includegraphics[width=7in]{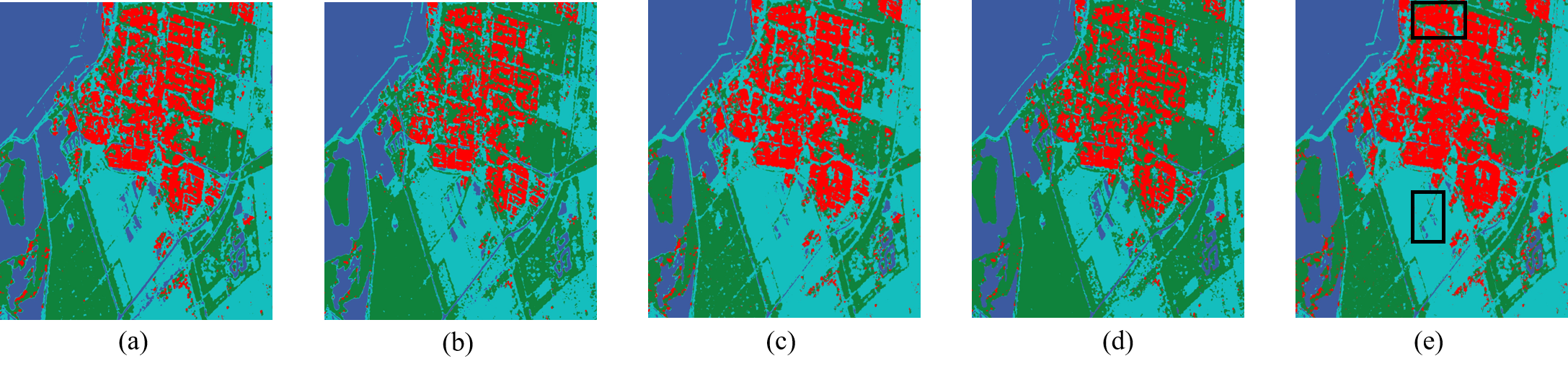}
	\captionsetup{font={small}, justification=raggedright}
	\caption{Few-shot classification results with different methods on the RADARSAT-2 Flevoland dataset. (a) MI-SSL. (b) CF-CSSL. (c) PCLNet. (d) SSPRL. (e) HCLNet.}
	\label{fig:fig9}
\end{figure*}

\begin{figure}[htb]
	\setlength{\abovecaptionskip}{-1mm}
	\setlength{\belowcaptionskip}{-3mm}
	\centering
	\includegraphics[width=3.5in]{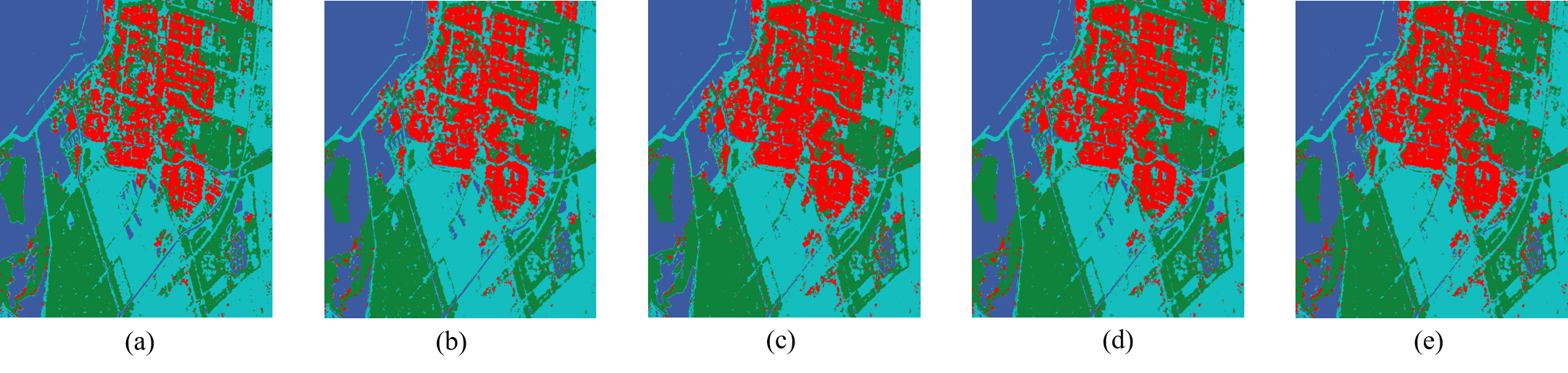}
	\captionsetup{font={small}, justification=raggedright}
	\caption{Full-sample classification results with different methods on the RADARSAT-2 Flevoland dataset. (a) MI-SSL. (b) CF-CSSL. (c) PCLNet. (d) SSPRL. (e) HCLNet.}
	\label{fig:fig10}
\end{figure}

\begin{figure}[htb]
	\setlength{\abovecaptionskip}{-1mm}
	\setlength{\belowcaptionskip}{-3mm}
	\centering
	\includegraphics[width=3.5in]{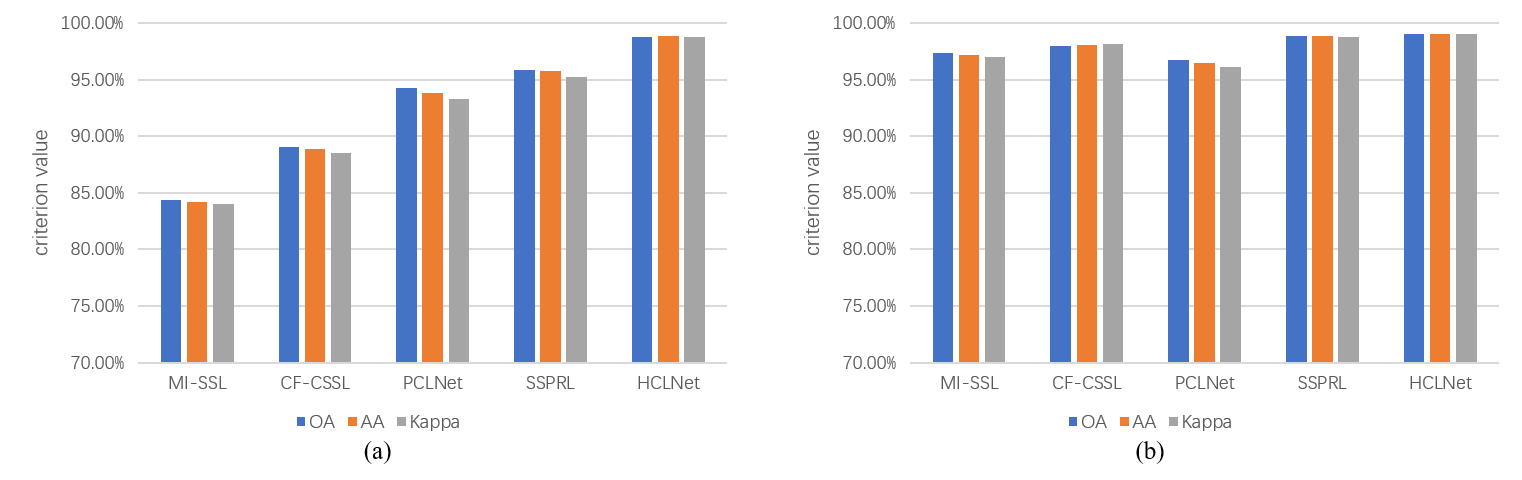}
	\captionsetup{font={small}, justification=raggedright}
	\caption{Comparisons of different methods on the AIRSAR Flevoland dataset. (a) Few-shot result. (b) Full-sample result.}
	\label{fig:fig11}
\end{figure}

\begin{figure*}[htb]
	\setlength{\abovecaptionskip}{-3mm}
	\setlength{\belowcaptionskip}{-3mm}
	\centering
	\includegraphics[width=7in]{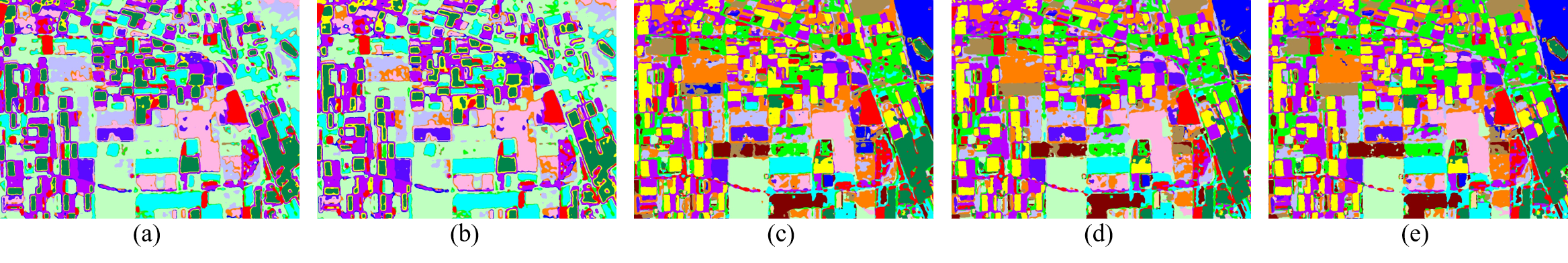}
	\captionsetup{font={small}, justification=raggedright}
	\caption{Few-shot classification results with different methods on the AIRSAR Flevoland dataset. (a) MI-SSL. (b) CF-CSSL. (c) PCLNet. (d) SSPRL. (e) HCLNet.}
	\label{fig:fig12}
\end{figure*}

\begin{figure}[htb]
	\setlength{\abovecaptionskip}{-1mm}
	\setlength{\belowcaptionskip}{-3mm}
	\centering
	\includegraphics[width=3.5in]{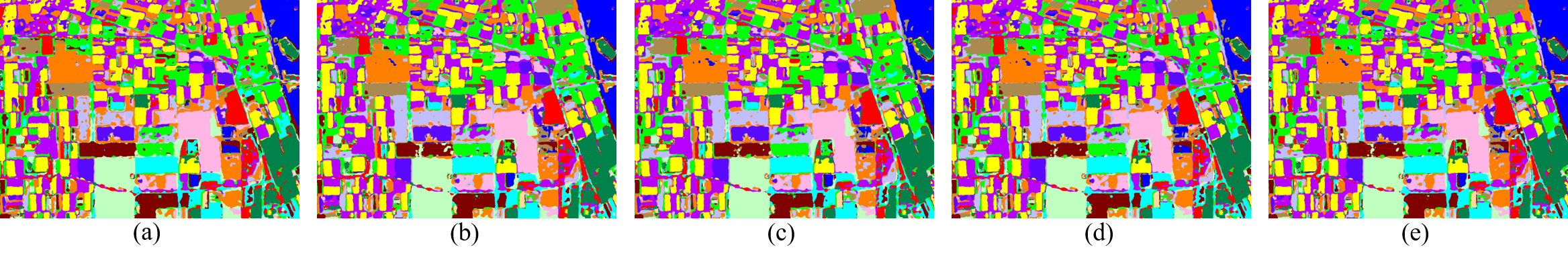}
	\captionsetup{font={small}, justification=raggedright}
	\caption{Full-sample classification results with different methods on the AIRSAR Flevoland dataset. (a) MI-SSL. (b) CF-CSSL. (c) PCLNet. (d) SSPRL. (e) HCLNet.}
	\label{fig:fig13}
\end{figure}

\begin{figure}[htb]
	\setlength{\abovecaptionskip}{-1mm}
	\setlength{\belowcaptionskip}{-3mm}
	\centering
	\includegraphics[width=3.5in]{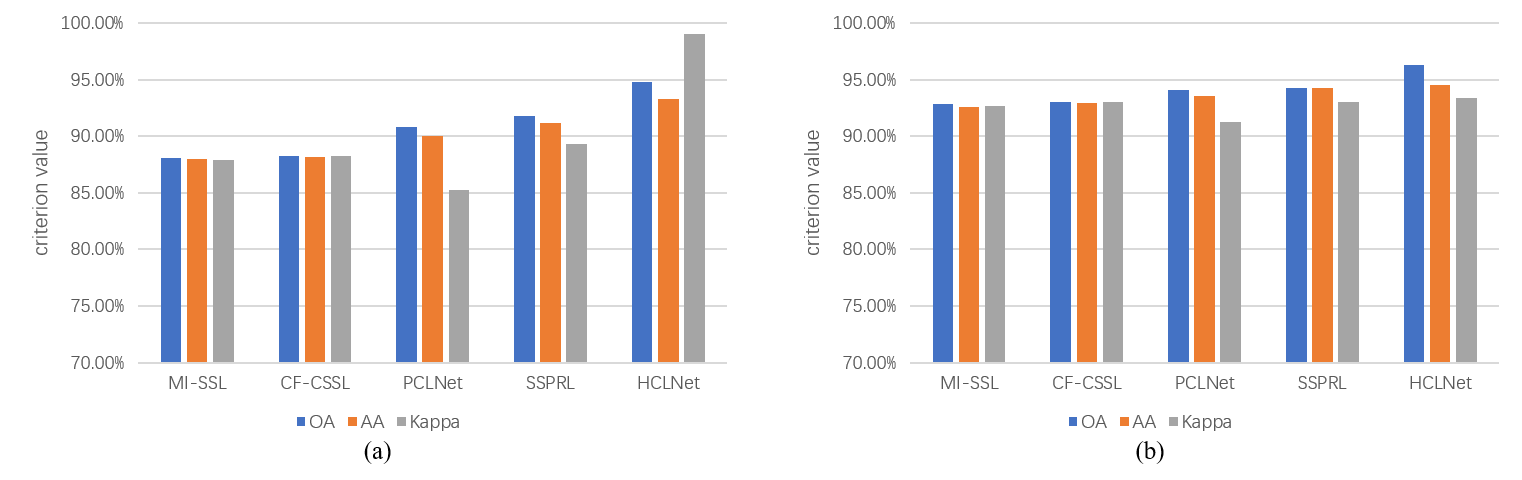}
	\captionsetup{font={small}}
	\caption{Comparisons of different methods on the ESAR Oberpfaffenhofen dataset. (a) Few-shot result. (b) Full-sample result.}
	\label{fig:fig14}
\end{figure}

\begin{figure*}[htb]
	\setlength{\abovecaptionskip}{-1mm}
	\setlength{\belowcaptionskip}{-3mm}
	\centering
	\includegraphics[width=7in]{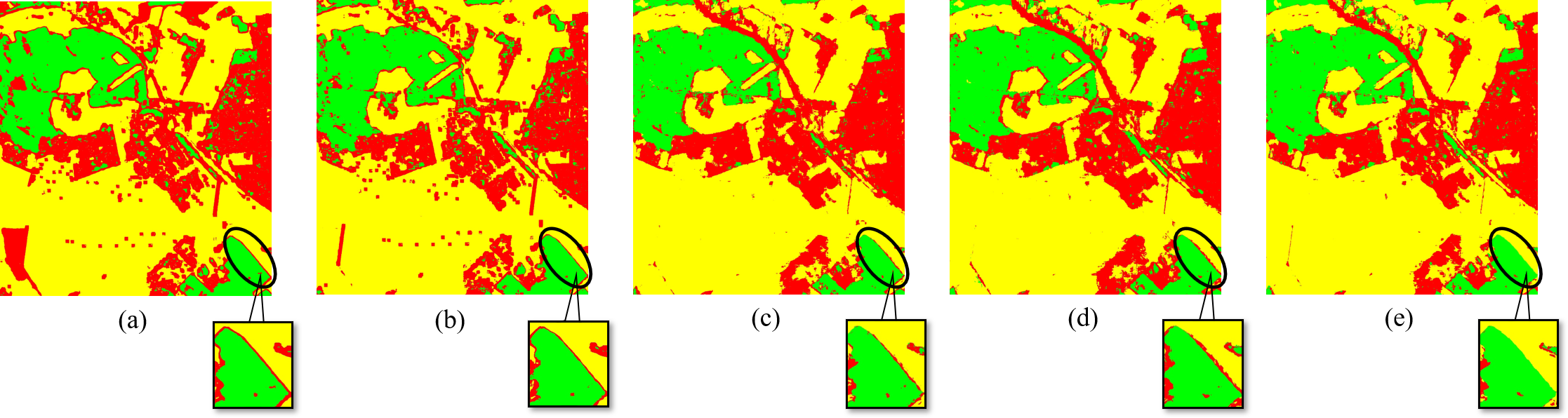}
	\captionsetup{font={small}, justification=raggedright}
	\caption{Few-shot classification results with different methods on the ESAR Oberpfaffenhofen dataset. (a) MI-SSL. (b) CF-CSSL. (c) PCLNet. (d) SSPRL. (e) HCLNet.}
	\label{fig:fig15}
\end{figure*}

\begin{figure}[htb]
	\setlength{\abovecaptionskip}{-1mm}
	\setlength{\belowcaptionskip}{-3mm}
	\centering
	\includegraphics[width=3.5in]{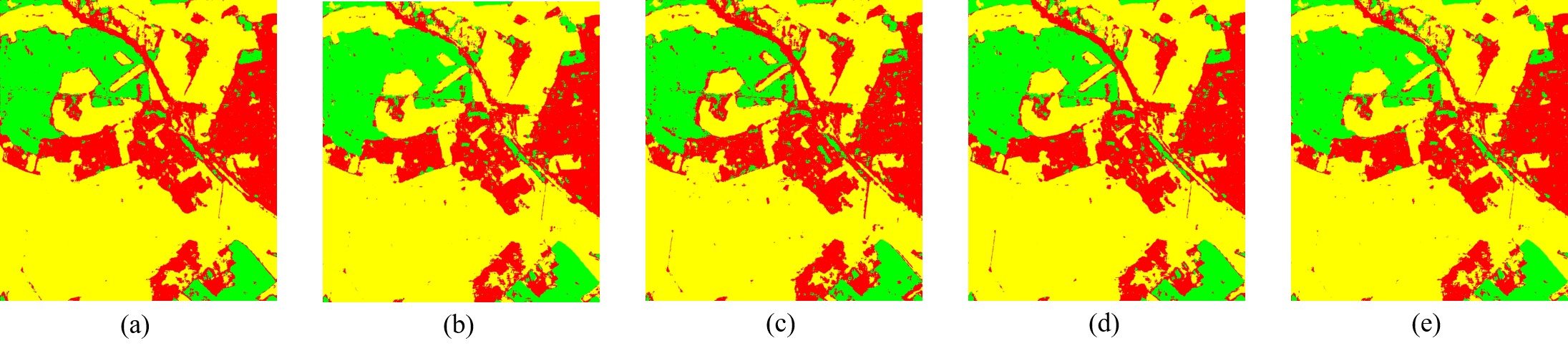}
	\captionsetup{font={small}, justification=raggedright}
	\caption{Full-sample classification results with different methods on the ESAR Oberpfaffenhofen dataset. (a) MI-SSL. (b) CF-CSSL. (c) PCLNet. (d) SSPRL. (e) HCLNet.}
	\label{fig:fig16}
\end{figure}

\begin{figure*}[htb]
	\setlength{\abovecaptionskip}{-3mm}
	\setlength{\belowcaptionskip}{-3mm}
	\centering
	\includegraphics[width=7in]{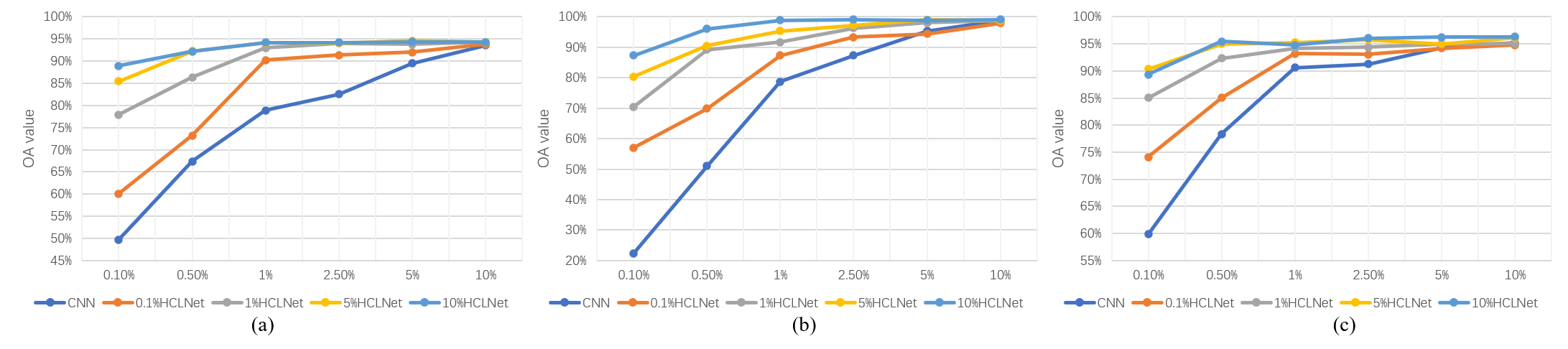}
	\captionsetup{font={small}, justification=raggedright}
	\caption{Comparisons of the performance(OA) with different ratios of unlabeled and labeled samples between CNN and HCLNet on three datasets. (a) RADARSAT-2 Flevoland. (b) AIRSAR Flevoland. (c) ESAR Oberpfaffenhofen.}
	\label{fig:fig17}
\end{figure*}

\begin{itemize}
	\item RADARSAT-2 Flevoland: As shown in Fig.\ref{fig:fig5}, a C-band, fully polarimetric image of the area of Netherland is obtained through the RADARSAT-2 system and was produced in April 2008. The size of the sub-image is $2375{\times}1635$. It identifies four types of ground objects: forest, farmland, city, and water area.

	\item AIRSAR Flevoland: An L-band, full polarimetric PolSAR image of the region of Flevoland, Netherlands, $750{\times}1024$, is obtained through the NASA/Jet Propulsion Laboratory AIRSAR. There are 15 labeled objects, including forest, rapeseed, beet, bare soil, grasses, peas, lucerne, barley,  buildings, potatoes, water, stembeans, and three kinds of wheat shown in Fig.\ref{fig:fig6}.

	\item ESAR Oberpfaffenhofen: It covers Oberpfaffenhofen, Germany, which is an L-band, full polarimetric image and is obtained through ESAR airborne platform. The size of the image is $1200{\times}1300$. Its ground-truth map is shown in Fig.\ref{fig:fig7}, which contains three classes: built-up areas, wood land, and open areas.
\end{itemize}

\begin{table}[htb]
	\setlength{\belowcaptionskip}{-3mm}
	\centering
	\captionsetup{font={small}}
	\caption{AllOF THE TARGET DECOPOSITION FEATURES}
	\label{table:tab1}
	\begin{tabular}{m{2cm}<{\centering}m{4.1cm}<{\centering}m{1cm}<{\centering}}
		\toprule[2pt]
		\textbf{Target Decomposition} & \textbf{Feature Name} & \textbf{Number} \\
		\midrule %[2pt]
		Krogager & sphere, diplane, helix & 3 \\
		TSVM & alpha-s, phi-s, phi, tau-m & 4 \\
		Neuman & delta, psi, tau & 3 \\
		Huynen & (T11,T22,T33)dB & 3 \\
		Holm & Holm1:(T11,T22,T33)dB, Holm2:(T11,T22,T33)dB & 6 \\
		Freeman & Freeman2:(Vol,Ground)dB, Freeman3:(Odd,Dbl,Vol)dB & 5 \\
		Cloude & (T11,T22,T33)dB & 3 \\
		Barnes & Barnes1:(T11, T22, T33)dB, Barnes2:(T11, T22, T33)dB & 6 \\
		ANNED & (Odd, Dbl, Vol)dB & 3 \\
		AnYang & AnYang3:(Odd,Dbl,Vol)dB, AnYang4:(Odd, Dbl,Vol, Hlx)dB & 7 \\
		H/A/$\alpha$ & alpha, anisotropy, beta, delta, entropy, gamma, lambda, combination: HA, (1-H)A, H(1-A), (1-H)(1-A) & 11 \\
		Yamaguchi & Yamaguchi3:(Odd, Dbl, Vol)dB, Yamaguchi4:(Odd, Dbl, Vol, Hlx)dB & 7 \\
		Vanzyl & (Odd, Dbl, Vol)dB & 3 \\
		MCSM & (Odd, Dbl, Vol, Hlx, Dbl-Hlx, Wire)dB & 6 \\
		\midrule %[2pt]
		\textbf{SUM} & {} & \textbf{70} \\
		\bottomrule[2pt]
	\end{tabular}
\end{table}

\begin{table}[htb]
	\setlength{\belowcaptionskip}{-3mm}
	\center
	\captionsetup{font={small}}
	\caption{FEW-SHOT CLASSIFICATION RESULTS (\%) ON THE RADARSAT-2 FLEVOLAND WITH DIFFERENT METHODS}
	\label{table:tab2}
	\begin{tabular}{m{1cm}<{\centering}m{1cm}<{\centering}m{1cm}<{\centering}m{1cm}<{\centering}m{1cm}<{\centering}m{1cm}<{\centering}}
		\toprule[2pt]
		\textbf{method} & MI-SSL & CF-CSSL & PCLNet & SSPRL & HCLNet \\
		\midrule
		\emph{Forest} & 79.23 & 69.70 & 86.55 & 72.35 & \textbf{91.69} \\
		\emph{Cropland} & 98.78 & 98.33 & 99.53 & 99.21 & \textbf{99.69} \\
		\emph{Water} & 93.02 & 96.75 & 92.06 & \textbf{98.36} & 93.87 \\
		\emph{Urban} & 91.10 & 85.35 & \textbf{95.60} & 77.10 & 94.85 \\
		\midrule
		\emph{OA} & 90.42 & 89.23 & 93.01 & 93.25 & \textbf{94.15} \\
		\emph{AA} & 89.54 & 88.64 & 92.00 & 92.50 & \textbf{93.74} \\
		\emph{Kappa} & 88.37 & 88.05 & 90.20 & 92.15 & \textbf{91.84} \\
		\bottomrule[2pt]
	\end{tabular}
\end{table}

\begin{table}[htb]
	\setlength{\belowcaptionskip}{-3mm}
	\center
	\captionsetup{font={small}}
	\caption{FULL-SAMPLE CLASSIFICATION RESULTS (\%) ON THE RADARSAT-2 FLEVOLAND WITH DIFFERENT METHODS}
	\label{table:tab3}
	\begin{tabular}{m{1cm}<{\centering}m{1cm}<{\centering}m{1cm}<{\centering}m{1cm}<{\centering}m{1cm}<{\centering}m{1cm}<{\centering}}
		\toprule[2pt]
		\textbf{method} & MI-SSL & CF-CSSL & PCLNet & SSPRL & HCLNet \\
		\midrule
		\emph{Forest} & 82.54 & 85.03 & 90.33 & 86.62 & \textbf{90.34} \\
		\emph{Cropland} & 97.98 & 97.28 & 97.90 & \textbf{98.68} & 98.36 \\
		\emph{Water} & 92.37 & 93.04 & 94.37 & \textbf{95.15} & 94.24 \\
		\emph{Urban} & 91.04 & 92.11 & 90.56 & 91.38 & \textbf{92.71} \\
		\midrule
		\emph{OA} & 90.95 & 92.05 & 93.29 & 93.70 & \textbf{94.30} \\
		\emph{AA} & 90.23 & 91.97 & 92.91 & 92.96 & \textbf{93.91} \\
		\emph{Kappa} & 89.72 & 91.80 & 91.06 & 91.17 & \textbf{92.04} \\
		\bottomrule[2pt]
	\end{tabular}
\end{table}

\subsection{Experimental Settings}
\label{section:4.2}
\begin{itemize}
	\item Implement details: The online network, in which the input patch size is $15{\times}15$, has two 2-D convolution layers in which the kernel size is $3{\times}3$, the padding is $2{\times}2$ and $1{\times}1$, and two linear embedding layers. The target network has two 1-D convolution layers in which the kernel size is the same as the online network, and the padding is $2{\times}2$, and one linear embedding layer. After each convolution layer, the batch norm and max pooling layers are added. Furthermore, the same as MoCo \cite{he2020momentum}, we use normalization in the outputs of both networks. The optimizer is the SGD with the momentum is $0.9$ and weight decay is $0.0001$, and the learning rate is initialized to $0.01$, which decreases with a cosine trend with training epoch. The $\tau$ is $0.07$. The network is trained for $30$ epochs and the minibatch size of $4096$. The threshold $\theta$ in the feature filter is set to $8$ obtained by cross validation, and the $k$ is set to $2$ in the first three rounds of search and $1$ in the subsequent searches. We initialize the number $K$ of superpixels based on the size of each superpixel being roughly $30{\times}30$. So the parameter $K$ for the three datasets are $1746$, $863$, and $1728$, respectively. The search range of the center of each superpixel is $3{\times}3$. All experiments were conducted independently on a single GeForce 3070 GPU with the PyTorch library. 
	
	\item Multi-features: The Refined Lee filter with the window size $7{\times}7$ is used to preprocess the three PolSAR datasets to reduce the influence of speckles on the result of classification. Then the same as \cite{yang2019cnn}, we use $14$ groups of target decomposition features and obtain $70$ decompositions features as the initial features of the feature filter. The whole target decomposition methods are shown in Table \ref{table:tab1}. These features are sufficient to represent PolSAR data.
	
	\item Compared methods: To evaluate the superiority of the proposed method, we select several semi-supervised and PolSAR-tailored CL methods for comparison. Specifically, four SOTA classification methods are chosen, including MI-SSL \cite{ren2021mutual}, Coarse-to-Fine CSSL (CF-CSSL) \cite{yang2022coarse}, PCLNet \cite{zhang2020unsupervised}, and SSPRL \cite{zhang2022exploring}.
\end{itemize}

\subsection{Experimental Results and Analysis}
\label{section:4.3}
\begin{table}[htb]
	\setlength{\belowcaptionskip}{-3mm}
	\center
	\captionsetup{font={small}}
	\caption{FEW-SHOT CLASSIFICATION RESULTS (\%) ON THE AIRSAR FLEVOLAND WITH DIFFERENT METHODS}
	\label{table:tab4}
	\begin{tabular}{m{1cm}<{\centering}m{1cm}<{\centering}m{1cm}<{\centering}m{1cm}<{\centering}m{1cm}<{\centering}m{1cm}<{\centering}}
		\toprule[2pt]
		\textbf{method} & MI-SSL & CF-CSSL & PCLNet & SSPRL & HCLNet \\
		\midrule
		\emph{Buildings} & 96.08 & 95.69 & 97.24 & 98.32 & \textbf{99.95} \\
		\emph{Rapeseed} & 53.92 & 37.23 & 87.32 & 90.60 & \textbf{96.91} \\
		\emph{Beet} & 87.23 & 92.03 & 98.33 & 99.02 & \textbf{99.77} \\
		\emph{Stembeans} & 83.10 & 69.15 & 98.18 & 98.89 & \textbf{99.93} \\
		\emph{Peas} & 86.54 & 85.73 & 95.24 & 96.32 & \textbf{99.21} \\
		\emph{Forest} & 85.01 & 67.23 & 96.38 & 97.93 & \textbf{95.11} \\
		\emph{Lucerne} & 91.99 & 92.19 & 95.02 & 97.28 & \textbf{99.13} \\
		\emph{Potatoes} & 93.12 & 93.02 & 92.37 & 95.00 & \textbf{99.84} \\
		\emph{Bare soil} & 94.13 & 92.95 & 89.88 & 92.13 & \textbf{98.52} \\
		\emph{Grass} & 89.26 & 83.72 & 85.42 & 89.26 & \textbf{98.47} \\
		\emph{Barley} & 92.19 & 92.43 & 95.77 & 96.23 & \textbf{97.75} \\
		\emph{Water} & 69.54 & 64.29 & 90.28 & 95.26 & \textbf{98.87} \\
		\emph{Wheat one} & 92.01 & 84.07 & 97.46 & 97.33 & \textbf{99.71} \\
		\emph{Wheat two} & 89.60 & 90.18 & 93.02 & 95.40 & \textbf{99.92} \\
		\emph{Wheat three} & 97.28 & 77.52 & 95.38 & 97.56 & \textbf{99.32} \\
		\midrule
		\emph{OA} & 84.32 & 89.02 & 94.30 & 95.83 & \textbf{98.80} \\
		\emph{AA} & 84.21 & 88.90 & 93.82 & 95.77 & \textbf{98.82} \\
		\emph{Kappa} & 83.98 & 88.54 & 93.27 & 95.23 & \textbf{98.73} \\
		\bottomrule[2pt]
	\end{tabular}
\end{table}

\begin{table}[htb]
	\setlength{\belowcaptionskip}{-3mm}
	\center
	\captionsetup{font={small}}
	\caption{FULL-SAMPLE CLASSIFICATION RESULTS (\%) ON THE AIRSAR FLEVOLAND WITH DIFFERENT METHODS}
	\label{table:tab5}
	\begin{tabular}{m{1cm}<{\centering}m{1cm}<{\centering}m{1cm}<{\centering}m{1cm}<{\centering}m{1cm}<{\centering}m{1cm}<{\centering}}
		\toprule[2pt]
		\textbf{method} & MI-SSL & CF-CSSL & PCLNet & SSPRL & HCLNet \\
		\midrule
		\emph{Buildings} & 99.07 & 99.53 & 98.30 & 99.53 & \textbf{99.91} \\
		\emph{Rapeseed} & 95.32 & 95.19 & 93.24 & 96.31 & \textbf{97.14} \\
		\emph{Beet} & 97.03 & 98.24 & 99.01 & 99.02 & \textbf{99.78} \\
		\emph{Stembeans} & 98.50 & \textbf{99.99} & 99.26 & 99.55 & 99.96 \\
		\emph{Peas} & 99.02 & 97.20 & 98.93 & 99.08 & \textbf{99.12} \\
		\emph{Forest} & 85.36 & 96.23 & \textbf{97.41} & 94.37 & 96.04 \\
		\emph{Lucerne} & 98.05 & 98.34 & 96.02 & 98.99 & \textbf{99.81} \\
		\emph{Potatoes} & 99.23 & 99.57 & 95.44 & \textbf{99.82} & 99.74 \\
		\emph{Bare soil} & 97.34 & 99.00 & 98.21 & 99.05 & \textbf{99.26} \\
		\emph{Grass} & 97.03 & 95.24 & 90.43 & 98.17 & \textbf{98.49} \\
		\emph{Barley} & 98.62 & 99.01 & 98.09 & \textbf{99.56} & 98.77 \\
		\emph{Water} & 95.44 & 98.98 & 91.52 & \textbf{99.79} & 98.84 \\
		\emph{Wheat one} & 98.79 & 99.26 & 97.40 & \textbf{99.93} & 99.69 \\
		\emph{Wheat two} & 99.91 & 99.90 & 95.88 & 99.84 & \textbf{99.95} \\
		\emph{Wheat three} & 97.36 & 96.51 & 98.31 & 99.40 & \textbf{99.52} \\
		\midrule
		\emph{OA} & 97.32 & 97.95 & 96.77 & 98.84 & \textbf{99.05} \\
		\emph{AA} & 97.14 & 98.02 & 96.50 & 98.83 & \textbf{99.06} \\
		\emph{Kappa} & 96.98 & 98.13 & 96.14 & 98.79 & \textbf{98.99} \\
		\bottomrule[2pt]
	\end{tabular}
\end{table}

\begin{table}[htb]
	\setlength{\belowcaptionskip}{-3mm}
	\center
	\captionsetup{font={small}}
	\caption{FEW-SHOT CLASSIFICATION RESULTS (\%) ON THE ESAR OBERPFAFFENHOFEN WITH DIFFERENT METHODS}
	\label{table:tab6}
	\begin{tabular}{m{1cm}<{\centering}m{1cm}<{\centering}m{1cm}<{\centering}m{1cm}<{\centering}m{1cm}<{\centering}m{1cm}<{\centering}}
		\toprule[2pt]
		\textbf{method} & MI-SSL & CF-CSSL & PCLNet & SSPRL & HCLNet \\
		\midrule
		\emph{Built-up areas} & 84.53 & 82.37 & 85.38 & \textbf{87.79} & 87.02 \\
		\emph{Wood land} & 89.22 & 91.05 & 89.70 & 90.88 & \textbf{94.50} \\
		\emph{Open areas} & 88.96 & 94.13 & 95.06 & 94.94 & \textbf{98.33} \\
		\midrule
		\emph{OA} & 88.05 & 88.23 & 90.83 & 91.78 & \textbf{94.80} \\
		\emph{AA} & 87.96 & 88.17 & 90.05 & 91.20 & \textbf{93.29} \\
		\emph{Kappa} & 87.89 & 88.25 & 85.21 & 89.32 & \textbf{92.99} \\
		\bottomrule[2pt]
	\end{tabular}
\end{table}

\begin{table}[htb]
	\setlength{\belowcaptionskip}{-3mm}
	\center
	\captionsetup{font={small}}
	\caption{FULL-SAMPLE CLASSIFICATION RESULTS (\%) ON THE ESAR OBERPFAFFENHOFEN WITH DIFFERENT METHODS}
	\label{table:tab7}
	\begin{tabular}{m{1cm}<{\centering}m{1cm}<{\centering}m{1cm}<{\centering}m{1cm}<{\centering}m{1cm}<{\centering}m{1cm}<{\centering}}
		\toprule[2pt]
		\textbf{method} & MI-SSL & CF-CSSL & PCLNet & SSPRL & HCLNet \\
		\midrule
		\emph{Built-up areas} & 88.35 & 85.63 & 89.08 & 90.18 & \textbf{90.34} \\
		\emph{Wood land} & 94.88 & 95.02 & \textbf{95.23} & 94.27 & 95.20 \\
		\emph{Open areas} & 95.63 & 96.04 & 96.30 & 97.35 & \textbf{97.92} \\
		\midrule
		\emph{OA} & 92.87 & 93.02 & 94.09 & 94.30 & \textbf{96.33} \\
		\emph{AA} & 92.56 & 92.95 & 93.54 & 94.27 & \textbf{94.49} \\
		\emph{Kappa} & 92.63 & 92.99 & 91.22 & 92.99 & \textbf{93.41} \\
		\bottomrule[2pt]
	\end{tabular}
\end{table}

\subsubsection{Classification accuracy}
\label{section:4.3.1}
In the experiment, our method is pre-trained with 10\% unlabeled data in each dataset. To verify the effectiveness of HCLNet, we choose 0.1\% and 10\% samples per category for training, denoted as few-shot and full-sample classifications. The rest of the samples are used as the test set for evaluation. The overall classification accuracy (OA), average accuracy (AA), and kappa coefficient (Kappa) are used as criteria to assess the performance of all methods. 

The results of the three datasets are shown in Tables \ref{table:tab2} to \ref{table:tab6}, respectively, demonstrating the superiority of HCLNet in the small number of labeled data. Different cases will generally have different results, but the trend is consistent across different datasets. 

\textbf{RADARSAT-2 Flevoland}: Specifically, in Tables \ref{table:tab2} and \ref{table:tab3}, for RADARSAT-2 Flevoland, MI-SSL is relatively stable, dropping only 0.53\%, 0.69\%, and 1.35\% from full-sample classification to few-shot classification, but its overall classification accuracy is not high and only 90.95\% in full-sample classification. CF-CSSL performs well in full-sample classification; however, its performance drops sharply, even lower than MI-SSL, when the number of data decreases, that is when few-shot classification. Numerically, from full-sample classification to few-shot classification, the OA, AA, and Kappa of CF-CSSL decrease by 2.82\%, 3.33\%, and 3.75\%. The overall number of network parameters of PCLNet is similar to that of HCLNet, which is 1.5 larger than HCLNet. Through the customized task and positive/negative sample selection of PCLNet, its accuracy is much improved compared with CF-CSSL, especially in few-shot classification, in which the overall improvements are 3.78\%, 3.36\%, and 2.15\%. However, HCLNet outperforms it in both full-sample classification and few-shot classification. In addition, due to the dual contrastive learning architecture of SSPRL and the unique selection of positive samples, its performance exceeds that of PCLNet. Numerically, the OA, AA, and Kappa of SSPRL are 0.24\%, 0.5\%, and 1.95\% higher than that of PCLNet, respectively. However, the network architecture of SSPRL is too complex, and the number of parameters is too large, which is more than ten times that of HCLNet. Furthermore, its performance fails to surpass HCLNet. The proposed HCLNet obtains the best results in full-sample and few-shots, in which the OA, AA, and Kappa are 94.15\%, 93.74\%, and 91.84\%.

For a more straightforward comparison, Fig.\ref{fig:fig8} illustrates the results of different methods in few-shot classification and full-sample classification of RADARSAT-2 Flevoland. The result clearly shows that HCLNet comprehensively outperforms all other methods. Furthermore, the classification maps of few-shot and full-sample classification results for different methods are presented in Figs. \ref{fig:fig9} and \ref{fig:fig10}. It can be observed that the HCLNet achieves the best result of the different landforms. Moreover, as indicated by the black box in Fig.\ref{fig:fig9}, we find many scattered, isolated pixel in the four compared methods. In comparison, our approach can solve this problem well due to the introduction of superpixels to ensure better contextual consistency in the phase of CL.

In order to more intutively show the advantage of HCLNet in few-shot classification, we respectively train HCLNet and CNN with the same architecture as online network of HCLNet according to different numbers of traning samples, and compare the accuray differences between them. Specifically, to explore the importance of the number of unlabeled and labeled samples, we designed comparison experiments between HCLNet and its backbone model using OA as the metric. We use the same online network of HCLNet without the projector head as the backbone model. Its parameters are initialized randomly. The backbone network and HCLNet are evaluated using 0.1\%, 0.5\%, 1\%, 2.5\%, 5\%, and 10\% samples per category. Moreover, for unlabeled samples, we set different ratios of 0.1\%, 1\%, 5\%, and 10\% to pre-training HCLNet. Compared to the results, we can find that the gap between the conventional CNN and HCLNet becomes larger with less labeled data, shown in Fig.\ref{fig:fig17}(a) clearly. HCLNet performs similarly to the backbone network with 1\% labeled data when trained with 0.1\% labeled data and 10\% unlabeled data, as shown in Fig.\ref{fig:fig17}(a). When the ratio of labeled data is over 1\%, the performance of HCLNet is better than the backbone network in any labeled data ratio. It fully demonstrates the effectiveness of HCLNet, which learns robust high-level representations from unlabeled data.

\textbf{AIRSAR Flevoland}: Similar experimental results for AIRSAR Flevoland are shown in Tables \ref{table:tab4} and \ref{table:tab5}; the predicted map is shown in Figs. \ref{fig:fig11} and \ref{fig:fig12}. Since this dataset has more categories and fewer data per class, the actual data amount of 1\% of training samples is small. It further widens the performance gap between HCLNet and other methods. Since there are only a few superpixels in each class, the difference between samples is tremendous, which makes the training of HCLNet more effective. The OA, AA, and Kappa of HCLNet are 2.97\%, 3.05\%, and 3.5\% higher than the best previous method, demonstrating that HCLNet has a greater advantage over PCLNet and SSPRL in this case. Visually, as shown in Fig.\ref{fig:fig13}, MI-SSL and CF-CSSL misclassified many data due to the lack of training labeled data. The performance is significantly degraded compared to the other methods. As mentioned, HCLNet reduces the scattered, isolated pixels to obtain better contextual consistency in the maps than the other four methods.

Furthermore, as shown in Fig.\ref{fig:fig17}(b), the performance of HCLNet with the ratio of labeled data is 0.5\%, and the ratio of unlabeled data is 10\% is better than backbone network with the ratio of labeled data is 5\%, and achieves the highest result. When only 0.1\% of training samples per category are used, the OA gap between the two methods is the largest, reaching 65.34\%. In contrast, the backbone network requires at least 2.5\% training samples per category to achieve the same result. It demonstrates the great generalization of the representation learned by HCLNet.

\textbf{ESAR Oberpfaffenhofen}: Compared with the first two datasets, ESAR Oberpfaffenhofen has fewer categories and a larger number of each category. It may result in similar data, but our method still works well in this case. As shown in Table \ref{table:tab6}, Table \ref{table:tab7}, and Fig.\ref{fig:fig14}, compared to the other two methods, which provide a different strategy to reduce the similarity between data, HCLNet is 3.97\%, 3.24\%, 7.78\% higher than PCLNet and 3.02\%, 2.09\%, 3.76\% higher than SSPRL in the few-shot. It demonstrates the effectiveness of superpixel-based instance discrimination to reduce the similarity between data. It still can keep an excellent contextual consistency shown in Figs. \ref{fig:fig15} and \ref{fig:fig16}. Moreover, due to the more labeled data, it can be observed that when the ratio of labeled data is 0.5\%, HCLNet shows excellent performance, outperforming the backbone network with any ratio of labeled data, as shown in Fig.\ref{fig:fig17}(c).

To sum up, the experimental results on three datasets can confirm the generalization and high classification accuarcy of HCLNet. 
% contextual consistency

\begin{figure*}[htb]
	\centering
	\setlength{\abovecaptionskip}{-3mm}
	\setlength{\belowcaptionskip}{-3mm}
	\includegraphics[width=7in]{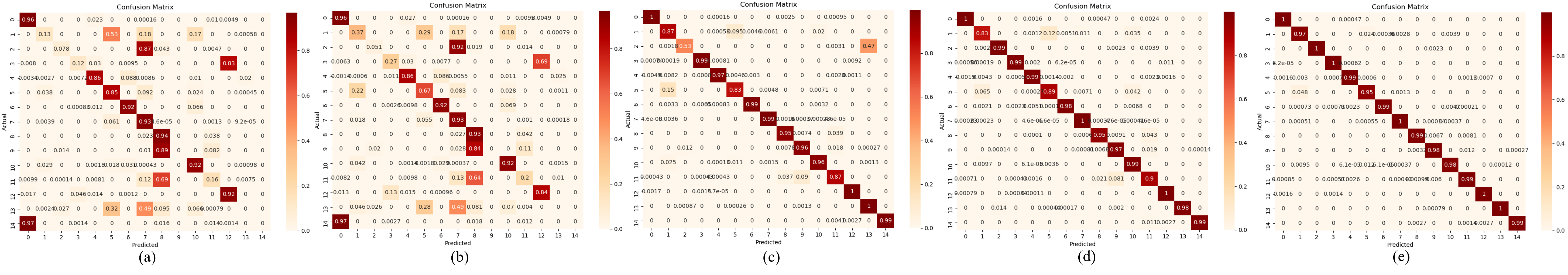}
	\captionsetup{font={small}}
	\caption{Confusion matrix of the classification result on the AIRSAR Flevoland dataset with different methods. (a) MI-SSL. (b) CF-CSSL. (c) PCLNet. (d) SSPRL. (e) HCLNet.}
	\label{fig:fig18}
\end{figure*}

\begin{figure}[htb]
	\centering
	\setlength{\abovecaptionskip}{0mm}
	\setlength{\belowcaptionskip}{-3mm}
	\includegraphics[width=3.5in]{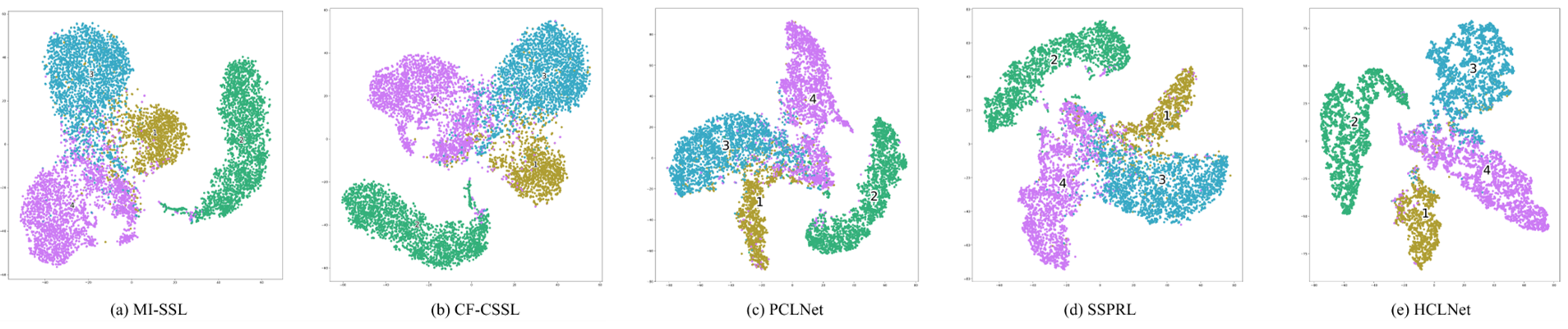}
	\captionsetup{font={small}}
	\caption{T-SNE visualization of the representations learned on the RADARSAT-2 Flevoland dataset with different methods. (a) MI-SSL. (b) CF-CSSL. (c) PCLNet. (d) SSPRL. (e) HCLNet.}
	\label{fig:fig19}
\end{figure}

\begin{figure}[htb]
	\centering
	\setlength{\abovecaptionskip}{0mm}
	\setlength{\belowcaptionskip}{-3mm}
	\includegraphics[width=3.5in]{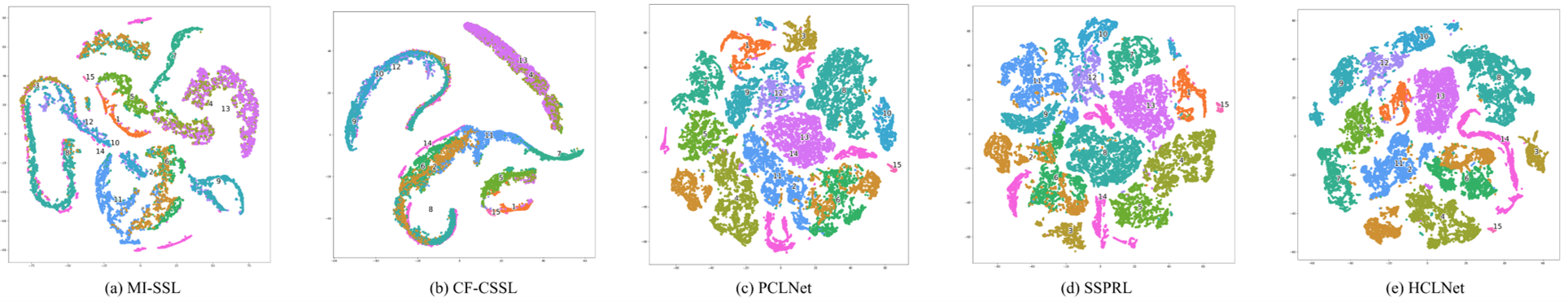}
	\captionsetup{font={small}}
	\caption{T-SNE visualization of the representations learned on the AIRSAR Flevoland dataset with different methods. (a) MI-SSL. (b) CF-CSSL. (c) PCLNet. (d) SSPRL. (e) HCLNet.}
	\label{fig:fig20}
\end{figure}

\begin{figure}[htb]
	\centering
	\setlength{\abovecaptionskip}{0mm}
	\setlength{\belowcaptionskip}{-3mm}
	\includegraphics[width=3.5in]{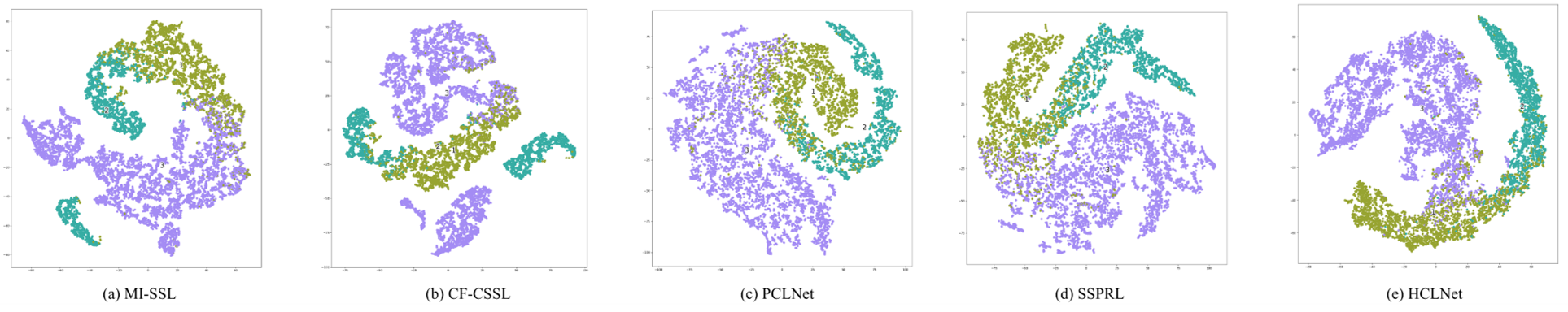}
	\captionsetup{font={small}}
	\caption{T-SNE visualization of the representations learned on the ESAR Oberpfaffenhofen dataset with different methods. (a) MI-SSL. (b) CF-CSSL. (c) PCLNet. (d) SSPRL. (e) HCLNet.}
	\label{fig:fig21}
\end{figure}

\begin{table*}[htb]
	\setlength{\belowcaptionskip}{-3mm}
	\center
	\captionsetup{font={small}}
	\caption{COMPARISON ON THE NUMBER OF PARAMETERS AND FLOPs FOR DIFFERENT METHODS.}
	\label{table:tab8}
	\begin{tabular}{m{4cm}<{\centering}m{1.7cm}<{\centering}m{1.7cm}<{\centering}m{1.7cm}<{\centering}m{1.7cm}<{\centering}m{1.7cm}<{\centering}m{1.7cm}<{\centering}}
		\toprule[2pt]
		Datasets & Complexity & MI-SSL & CF-CSSL & PCLNet & SSPRL & HCLNet \\
		\midrule
		RADARSAT-2 Flevoland & Parameter(M) & 1.67 & 12.35 & 0.34 & 5.79 & \textbf{0.28} \\
		RADARSAT-2 Flevoland & FLOPs(M) & 12.53 & 4.67 & 2.90 & 14.85 & \textbf{2.43} \\
		\midrule
		AIRSAR Flevoland & Parameter(M) & 2.09 & 12.32 & 0.59 & 8.63 & \textbf{0.28} \\
		AIRSAR Flevoland & FLOPs(M) & 20.63 & 7.96 & 5.12 & 25.39 & \textbf{4.74} \\
		\midrule
		ESAR Oberpfaffenhofen & Parameter(M) & 1.44 & 12.33 & 0.32 & 5.26 & \textbf{0.28} \\
		ESAR Oberpfaffenhofen & FLOPs(M) & 9.89 & 3.97 & 2.13 & 11.92 & \textbf{1.85} \\
		\bottomrule[2pt]
	\end{tabular}
\end{table*}

\begin{table}[htb]
	\setlength{\belowcaptionskip}{-3mm}
	\centering
	\caption{OA (\%) ON RADARSAT-2 FLEVOLAND ($OA_{RF}$), AIRSAR FLEVOLAND ($OA_{AF}$) AND ESAR OBERPFAFFENHOFEN ($OA_{EO}$) DATA SETS FOR ABLATION STUDY}
	\label{table:tab9}
	\begin{tabular}{m{2cm}<{\centering}m{1.7cm}<{\centering}m{1.7cm}<{\centering}m{1.7cm}<{\centering}}
		\toprule[2pt]
		\textbf{method} & $OA_{RF}$ & $OA_{AF}$ & $OA_{EO}$ \\
		\midrule
		\emph{CNN+SID+FF} & 90.22 & 95.35 & 91.37 \\
		\emph{Hnet+SID+FF} & 94.15 & 98.80 & 94.80 \\
		\emph{Hnet+SID} & 91.47 & 93.65 & 92.91 \\
		\emph{Hnet+FF} & 64.50 & 85.32 & 79.66 \\
		\bottomrule[2pt]
	\end{tabular}
\end{table}

\subsubsection{Scattering Confusion}
\label{section:4.3.2}
The degree to alleviate the scattering confusion problem is also the main evaluation of the effectiveness of the PolSAR classification method. So we utilize the \textbf{confusion matrix} and the \textbf{local classification visualization} to make a quantitative and qualitative analysis on the performance of HCLNet in scattering confusion problem. Specifically, we choose AIRSAR Flevoland dataset for the main experimental results presentation, mainly due to its large number of land cover categories, each land cover is more prone to scattering confusion. As show in Fig.\ref{fig:fig18}, it can be seen that the scattering confusion between different land cover classifications of MI-SSL and CF-CSSL is very serious. For example, in MI-SSL, there is Rapeseed certain scattering similarity between Forest, Potatoes, and Barley, which leads to the misclassification of these types of land covers by the model. And in CF-CSSL, besides having similar phenomenon to MISSL, it also has the scattering confusion problem between Forest and Potatoes leading to model classification errors. The situation of PCLNet and SSPRL is somewhat better, but there is still a partial scattering confusion problem. PCLNet still has some shortcomings in dealing with the scattering similarity of Beet and Wheat two. It cannot fully learn the difference between their scattering features, which leads to the model misclassifying part of Beet. Similarly, SSPRL does not fully learn the scattering difference between each land cover, which leads to some confusion in its classification of Rapeseed and Forest. However, HCLNet learns the similarity and difference between different instances well through the target network’s learning of scattering similarity and the online network’s learning of scattering difference. There are very few instances where a class is significantly misclassified to another, that is, there is no correlation between the classes. This shows that HCLNet has a great improvement in alleviating the scattering confusion problem.

Because of the scattering interference between adjacent boundaries of different land covers, the scattering confusion at the boundary is the most severe. Therefore we selected a typical land cover boundary line in the ESAR Oberpfaffenhofen dataset and showed the specific classification results of different methods. Specifically, we zoom in on the lower right part of each method's classification map, which is the boundary between Wood Land and Open Areas. As shown in the lower part of Fig.\ref{fig:fig15} , due to the lack of sufficient sample learning and the scattering interference, the first four methods all misclassify the boundary severely, resulting in “circles” in the classification map. In contrast, HCLNet basically classifies all the boundary pixels correctly. This graphically demonstrates the superiority of our method in mitigating the scattering confusion problem.

\subsubsection{Representation visualization}
\label{section:4.3.3}
In the above experiments, HCLNet has demonstrated the powerful ability of representation learning with unlabeled data using supervised-based instance discrimination. In order to further explore the quality of representation, 2-D t-stochastic neighbor embedding(t-SNE) \cite{van2008visualizing} is used to visualize the learned representation. Specifically, MI-SSL, CF-CSSL, PCLNet, SSPRL, and HCLNet utilize 1\% labeled samples per category in the map. For each method, the output of the representation encoder is used as the representation without any labeled samples. These visualizations are performed on three benchmark datasets above. The results are shown in Figs. \ref{fig:fig19}, \ref{fig:fig20}, and \ref{fig:fig21}, different colors indicate different categories. 

From the results, it can be observed that MI-SSL cannot distinguish each category well, with existing multiple categories highly overlapping. Moreover, the closeness is relatively weak. Some features of the same category are distributed in different locations and form multiple disconnected regions.

For CF-CSSL, under the guidance of only a few labeled data, it leads to the poor improvement of closeness overlapping, and some even deteriorate. Benefiting from training on large amounts of unlabeled data, PCLNet and SSPRL drastically reduce the overlapping but still exist some disconnected regions. On the contrary, HCLNet significantly alleviates disconnection and overlapping. It turns out that HCLNet provides a better representation by learning from different perspectives for the downstream task to improve the classification ability.

Through t-SNE, it also again illustrates the significant improvement of HCLNet on the scattering confusion problem. The representations obtained by HCLNet distinguish each land cover well, and there are few cases of severe overlap of representations in some classes as in other methods.

\subsubsection{Model Complexity Analysis}
\label{section:4.3.4}
We evaluated the complexity of the models using the number of model parameters and floating-point operations (FLOPs). The quantitative comparison results of all methods are listed in Table \ref{table:tab8}. All the compared methods follow the setup in the original paper. For instance, CF-CSSL use the the U-Net that follows an encoder-decoder architecture, which the encoder includes $3$ convolutional blocks, each of which consists of two $3{\times}3$ convolutional layers and one $2{\times}2$ maxpooling layer, and the decoder has a symmetric architecture to the encoder, which each block is composed of one up-sampling layer and three convolutional layers. It has a high number of parameters and FLOPs. MI-SSL has multiple forward computations and multiple contrast computations, so it has high FLOPs. PCLNet and SSPRL have higher FLOPs due to the complex auxiliary modules in their networks. In contrast, our method has a relatively low number of parameters and FLOPs because our network only has fewer layers, and not too many reduntant, complex designs.

\subsubsection{Ablation study}
\label{section:4.3.5}
To better understand the effectiveness of each component of HCLNet, we experiment with combinations of different components on the three datasets above using the setting of few-shot and design the three groups of ablation experiments. The ablation results are reported in Table \ref{table:tab9}. The abbreviations in Table \ref{table:tab9} represent the different components: CNN is the Siamese network in which the target network and the online network are the same network architecture, and both of them share parameters; Hnet is the heterogeneous network with the same architecture as HCLNet; SID is the superpixel-based Instance Discrimination; FF is the feature filter. The results implicate that each component can improve the classification results. The first two experiments involve that compared with CNN as the architecture, Hnet exhibits the more powerful representation extraction ability and has fewer parameters. When FF is added to remove the redundancy between features, the model can learn better high-level representation, demonstrated in the experiment's third group.

\begin{figure}[htb]
	\setlength{\abovecaptionskip}{-3mm}
	\setlength{\belowcaptionskip}{-3mm}
	\centering
	\includegraphics[width=3.5in]{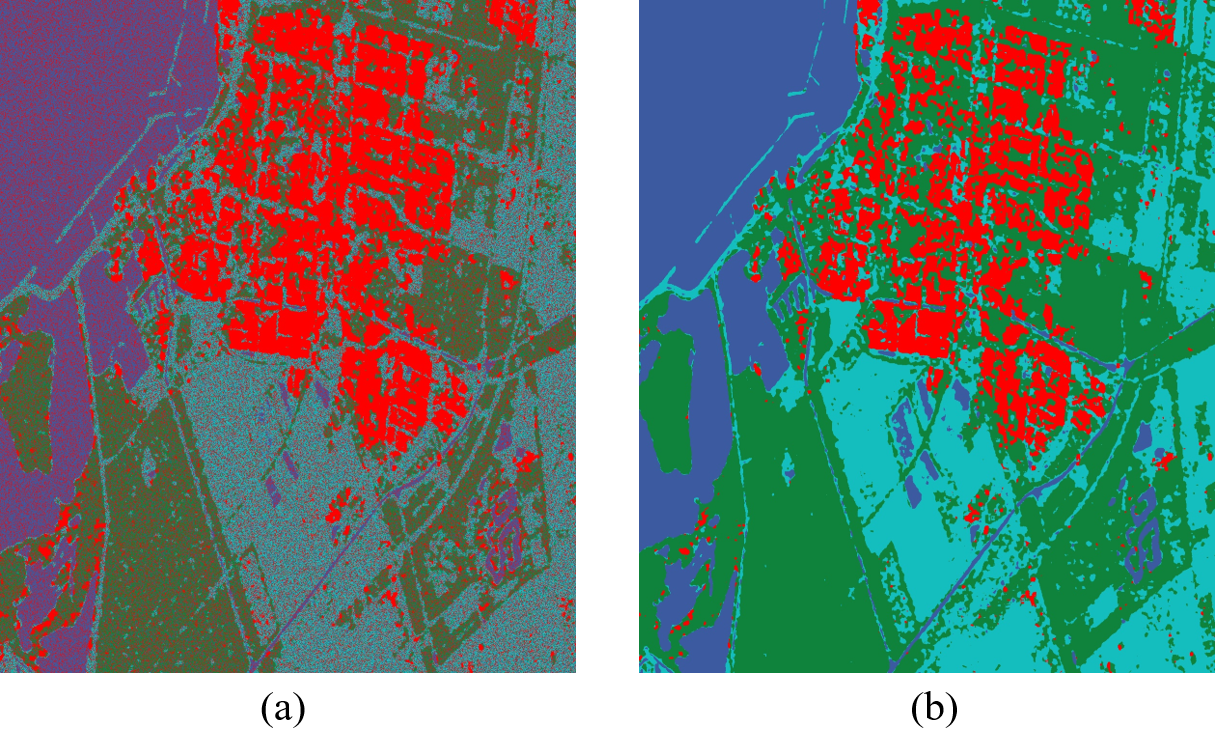}
	\captionsetup{font={small}}
	\caption{(a) The salt and pepper noise problem and (b) the improved result.}
	\label{fig:fig22}
\end{figure}

Significantly, the fourth group of experiments demonstrates that the SID is essential for heterogeneous networks. As shown in Fig.\ref{fig:fig22}(a), without SID, the training of HCLNet likely falls into the salt and pepper noise problem. We analyze that the network is challenging to learn the high-level representation when CL without SID to distinguish the differences between pixels and even confuse the representations between pixels of different categories. To verify our conjecture, we replace the SID and use pixels of different categories as negative samples and pixels of the same category as positive samples to increase the dissimilarity between negative samples. Finally, it successfully solves the salt and pepper noise problem, as shown in Fig.\ref{fig:fig21}(b). So starting from the direction of SSL, Hnet, and SID may be a match made in heaven. 

\section{Conclusion}
\label{section:5}
This article proposes a self-supervised learning method based on CL with a heterogeneous network for the first time. We use HCLNet to extract the high-level representation of PolSAR data from the physical and statistical features and propose two plugins to improve learning. The feature filter is introduced to select the appropriate combination of features to reduce the redundancy of multi-target decomposition features. The superpixel-based Instance Discrimination is proposed to reduce the similarity between pixels and learn better representations. Therefore, with the help of unsupervised pre-training to learn representation, the online network can achieve high results of few-shot PolSAR classification by fine-tuning. Experiments are conducted on three widely used benchmark datasets, and the experimental results demonstrate the performance of HCLNet compared with several mainstream methods in both few-shot and full-sample classification.

Compared with the CL of optical images, PolSAR has more valuable features under different properties, while the heterogeneous network has the natural advantage of entirely using these features. This work creates a precedent for future research on heterogeneous network learning. And we believe that more in-depth and comprehensive research about the heterogeneous network may further improve the PolSAR classifier performance. Our future interest is to explore the problem of positive and negative selection for heterogeneous networks in depth. 

\bibliographystyle{IEEEtran}
\nocite{*}
\bibliography{reference}

\end{document}